\documentclass{article} % For LaTeX2e
\usepackage{colm2024_conference}

\usepackage{booktabs}
\usepackage{graphicx}
\usepackage{enumitem}
\usepackage{wrapfig}
\usepackage{algorithm}
\usepackage{algpseudocode}
\usepackage{natbib}
\usepackage{makecell}
\usepackage{booktabs}
\usepackage{array}
\usepackage{amsmath}
\usepackage{amssymb}
\usepackage{amsfonts}
\usepackage{multirow}
\usepackage{verbatim}
\usepackage{caption}
\usepackage{longtable}
\usepackage{supertabular}
\usepackage{CJKutf8}
\usepackage[utf8]{inputenc} % optional
\usepackage[T1]{fontenc}
\usepackage[french,vietnamese,mongolian,greek,english]{babel}
\usepackage{pifont}

\usepackage{enumitem}
\usepackage{tablefootnote}
\usepackage{xspace}
\usepackage{textcomp}
\usepackage{makecell}
\usepackage{lscape} 
\usepackage{siunitx}
\usepackage{listings}
\usepackage{xcolor}
\usepackage{svg}
\definecolor{dt}{gray}{0.7}
\usepackage{pifont}       % \ding{xx}
\usepackage{bbding}       % \Checkmark,\XSolid,... (需要和pifont宏包共同使用)
\usepackage{fontawesome}

\usepackage{scrextend}

\usepackage{tgpagella}
\usepackage{latexsym}
\usepackage[T1]{fontenc}
\usepackage[utf8]{inputenc}
\usepackage{microtype}
\definecolor{mydarkblue}{rgb}{0,0.08,0.45}
\definecolor{citecolor}{HTML}{0071BC}
\usepackage{url}            % simple URL typesetting
\usepackage{nicefrac}       % compact symbols for 1/2, etc.
\usepackage{changepage}
\usepackage{xargs}          % Use more than one optional parameter in a new commands
\usepackage{wrapfig,lipsum,booktabs}
\usepackage{longtable}
\usepackage{subcaption}
\usepackage{endnotes}

\usepackage{pgfplots}
\usetikzlibrary{pgfplots.groupplots,fit,backgrounds}
\pgfplotsset{compat=1.3}
\usepackage{tikz}
\usetikzlibrary{patterns}

\usepackage[most]{tcolorbox}

\usepackage{hyperref}
\definecolor{darkblue}{rgb}{0, 0, 0.5}
\hypersetup{colorlinks=true, citecolor=darkblue, linkcolor=darkblue, urlcolor=darkblue, bookmarks=true, bookmarksopen=false, bookmarksnumbered=true}

\usepackage[capitalize,noabbrev]{cleveref}

\crefname{section}{\S}{\S\S}
\Crefname{section}{\S}{\S\S}
\crefname{subsection}{\S\S}{\S\S}
\Crefname{subsection}{\S\S}{\S\S}
\crefformat{section}{\S#2#1#3}
\crefformat{subsection}{\S\S#2#1#3}

\crefname{table}{Table}{Tables}
\crefname{figure}{Figure}{Figures}
\crefname{algorithm}{Algorithm}{}
\crefname{equation}{eq.}{}
\crefname{appendix}{Appendix}{}
\crefformat{section}{Section #2#1#3}
\usepackage{multicol}
\usepackage{tcolorbox}
\usepackage{titlesec}
\titleformat*{\section}{\large\bfseries}
\usepackage{array} % 导言区
\newcolumntype{P}[1]{>{\centering\arraybackslash}p{#1}} % 定义一个居中对齐的固定宽度列
\usepackage{multirow}

\usepackage[utf8]{inputenc} % Ensure UTF-8 encoding if you have non-ASCII characters, though not strictly needed for this English text
\newcolumntype{L}[1]{>{\raggedright\let\newline\\\arraybackslash\hspace{0pt}}p{#1}}
\newcolumntype{C}[1]{>{\centering\let\newline\\\arraybackslash\hspace{0pt}}p{#1}}
\usepackage{tikz}
\usetikzlibrary{backgrounds, fit}  % 加载需要的库
\usepackage{adjustbox}
\definecolor{objblue}{RGB}{3,139,221}  
\definecolor{attrred}{RGB}{255,67,67}    
\definecolor{easygreen}{RGB}{0,156,75}  
\definecolor{middleyellow}{RGB}{242,89,34}  
\definecolor{hardred}{RGB}{216,56,58}
\usepackage{colortbl}
\usepackage{array}

\usepackage[most]{tcolorbox}
\definecolor{BoxBackground}{RGB}{240, 240, 240} % 浅灰色背景
\definecolor{BoxFrame}{RGB}{0, 0, 0} % 黑色边框
\definecolor{TitleBackground}{RGB}{0, 0, 0} % 标题背景颜色
\definecolor{TitleText}{RGB}{255, 255, 255} % 标题文字颜色
\tcbset{
  academicbox/.style={
    boxsep=5pt,
    left=2pt,
    right=2pt,
    bottom=0.5pt,
    boxrule=0.5pt,
    colback=BoxBackground,
    colframe=BoxFrame,
    colbacktitle=TitleBackground,
    coltitle=TitleText,
    enhanced,
    attach boxed title to top left={yshift=-0.1in,xshift=0.1in},
    boxed title style={boxrule=0pt,colframe=white},
    title={#1},
  }
}
\newtcolorbox{AcademicBox}[1][]{academicbox=#1}

\title{Qwen-RobotWorld Technical Report: Unifying Embodied World Modeling through Language-Conditioned Video Generation}

\author{
\bf Qwen Team}

\begin{document}

\maketitle
\vspace{-1mm}
\begin{abstract}
\vspace{-4mm}
We introduce \textbf{\textsc{Qwen-RobotWorld}}, a language-conditioned video world model for embodied intelligence. With natural language as a unified action interface, it predicts physically grounded future visual trajectories from current observations across robotic manipulation, autonomous driving, indoor navigation, and human-to-robot transfer. This unified formulation provides three promising application directions: synthetic data generation for policy training augmentation, scalable virtual environments for policy evaluation, and language-guided planning signals for downstream robot control. This is achieved through a three-part design: a) \textbf{Double-Stream MMDiT with MLLM Action Encoding}, where a 60-layer double-stream diffusion transformer couples frozen Qwen2.5-VL semantics with video-VAE latents through layer-wise joint attention; b) \textbf{Embodied World Knowledge (EWK)}, an 8.6M video-text corpus (200M+ frames) with action-language mapping over 20+ embodiments and 500+ action categories; and c) \textbf{General+Expert Progressive Curriculum}, a two-stage training strategy that first learns general visual priors and then injects embodied specialization under a shared language interface. Extensive results show strong competitiveness: ranks 1st overall on EWMBench and DreamGen Bench, outperforms all open-source models on WorldModelBench and PBench. Additional zero-shot analyses on RoboTwin-IF benchmark further support robust generalization and multi-view consistency.
\begin{figure}[h]
    \centering
    \includegraphics[width=\linewidth]{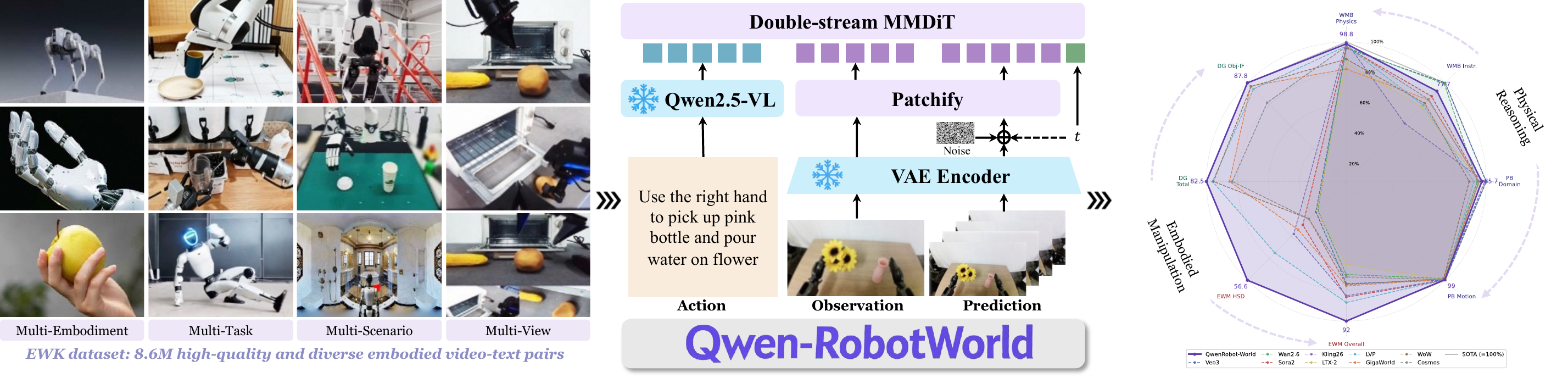}
    \label{fig:radar}
\end{figure}
\vspace{-8mm}
\end{abstract}

\section{Introduction}

Embodied intelligence requires agents to perceive, reason, and act within physical environments—spanning robotic manipulation at tabletop scale, autonomous navigation through urban traffic, and wayfinding across indoor spaces. Training such systems directly in the real world is costly, inefficient, and fraught with safety risks. World models offer a scalable alternative: by learning environment dynamics from observational data, they serve as interactive training platforms that allow embodied agents to acquire and refine behaviors without physical deployment.

A world model can be formalized as a \textit{state transition function}: given a current state $s_t$ and an action $a_t$, it predicts the resulting state $s_{t+1} = f(s_t, a_t)$~\cite{ye2026worldactionmodelszeroshot}. In video-based world models, states are visual observations (video frames or their latent representations), and the model generates future visual trajectories conditioned on the current observation and an action signal. The action $a_t$ can take various forms—low-level motor commands, high-level waypoint trajectories, or natural language instructions. Among these, \textbf{natural language is the most general and accessible action representation}~\cite{ye2026worldactionmodelszeroshot}: a single instruction such as ``pick up the red cup and place it on the shelf'' implicitly encodes the complete action sequence, goal state, and physical constraints, without requiring robot-specific control interfaces. Language actions can furthermore be utilized in two complementary directions: as an explicit \textit{input} fused into the model's condition signal to govern state transitions, or as an \textit{output} inferred post-hoc from generated video to serve as an action label. This flexibility positions language-conditioned world models as universal simulation backbones that generalize across embodied platforms without interface redesign.

However, a fundamental tension currently limits world model effectiveness. \textit{General video generation models}~\cite{sora2,veo3} learn rich visual priors from internet-scale data but fail to accurately model embodied physics—contact dynamics, rigid-body structural constraints, and action-consequence relationships that are critical for physically plausible state transitions. \textit{Domain-specific embodied models}~\cite{cosmos,lvp,gigaworld}, conversely, are tailored to individual scenarios (e.g., tabletop manipulation or driving); they rely on structured, robot-specific action representations such as joint angles or waypoints, which cannot generalize across embodiment types or task categories, fundamentally limiting their utility as cross-platform simulation environments.

Bridging this gap requires grounding diverse embodied experiences in general visual priors, with natural language as the unified action interface that enables cross-scenario and cross-task integration. Different embodied domains provide complementary physical knowledge that collectively enriches the world model's state transition function: \textit{manipulation} teaches fine-grained contact physics and object-state transformations within confined workspaces; \textit{autonomous driving} teaches large-scale multi-agent dynamics and 3D scene geometry through ego-motion parallax and scene-scale transitions; \textit{indoor navigation} teaches room-scale spatial reasoning, where language instructions must be grounded into spatially coherent visual trajectories over extended horizons. Because these domains share a common language interface, they can be trained jointly—with each domain's physical knowledge reinforcing the others rather than conflicting. Furthermore, translating human demonstrations into robot executions through video editing opens a practical pathway to scale embodied training data beyond the limits of physical robot collection.

We present \textbf{\textsc{Qwen-RobotWorld}}, a language-conditioned video world model in the Qwen series that realizes this vision through tightly coupled innovations in architecture, data, and training. Beyond high-fidelity action-conditioned prediction, the model serves as a unified backbone that, with task-specific adaptation, can support three representative embodied world model applications: a \textit{synthetic data engine}, a \textit{policy evaluation environment}, and an \textit{action planner}.

\textbf{Architecture: Double-Stream MMDiT with MLLM Action Encoding (\S\ref{sec:model}).}
To implement language-conditioned state transitions, we adopt a double-stream Multimodal Diffusion Transformer (MMDiT) backbone. An \textit{understanding stream} processes rich semantic features extracted by a frozen Qwen2.5-VL encoder, representing the action $a_t$; a \textit{generation stream} processes visual latents from a video-compatible VAE, representing the visual state $s_t$. The two streams interact via joint attention at every layer, enabling bidirectional cross-modal fusion throughout the denoising process. Using an MLLM as the action encoder—rather than lightweight encoders such as T5~\cite{raffel2020t5} or CLIP~\cite{clip}—yields two key advantages: (1) its deep language understanding accurately parses complex, compositional instructions into precise condition signals that govern fine-grained state transitions; (2) its internalized world knowledge (e.g., that robot arms are rigid bodies with fixed link lengths and joint constraints) implicitly constrains the space of physically plausible transitions, and—combined with T2I co-training—prevents object deformation across video frames without requiring explicit geometric prompts, a common failure mode in models lacking such semantic grounding.

\textbf{Data: Embodied World Knowledge Dataset (\S\ref{sec:data}).}
To train a state transition function that generalizes across embodied domains, we construct the Embodied World Knowledge (EWK) dataset—approximately 8.6M video-text pairs comprising over 200M observation frames. The corpus spans four embodied domains alongside general video data (30\% of the total): \textbf{manipulation} ($\sim$5.9M samples, 20+ robot morphologies, 1300+ skills) provides the core embodied foundation; \textbf{autonomous driving} ($\sim$200K samples from Waymo, NVIDIA PhysicalAI-AD, Bench2Drive, and Sekai) contributes large-scale ego-motion and multi-agent dynamics; \textbf{indoor navigation} (6K+ language-guided episodes from VLNVerse) provides room-scale spatial reasoning grounded in continuous trajectories; and \textbf{human-to-robot transfer} data—generated via an automated MANO~\cite{romero2017mano}-to-robot pipeline across 14 robot morphologies—enables cross-embodiment video editing. A central methodological contribution is our \textit{action-language mapping framework}, which standardizes actions across 20+ robot embodiment types and 500+ action categories into a unified natural language interface, yielding approximately 8.6M high-quality cross-scenario, cross-task embodied video-text pairs. This is complemented by task-aware temporal segmentation (ensuring each sample captures a complete, well-defined state transition) and a hierarchical five-layer viewpoint-aware annotation pipeline that substantially improves caption specificity and downstream instruction-following.

\textbf{Training: From General Priors to Embodied Specialization (\S\ref{sec:training}).}
We adopt a two-stage progressive training curriculum. In pretraining, joint training across T2I, T2V, and TI2V tasks over general-domain data builds foundational visual priors, with T2I specifically anchoring geometrically correct object morphology that transfers to video generation. In the SFT stage, embodied data is introduced progressively (70\% embodied, 30\% general) through a four-phase mixing schedule: single-view manipulation $\rightarrow$ multi-view expansion $\rightarrow$ multi-view concatenated generation $\rightarrow$ complex tasks and cross-domain data. Within the embodied portion, manipulation dominates at $\sim$90\% sampling weight to ensure depth of physical grounding, while multi-view concatenation and navigation/driving data each receive $\sim$5\% to provide breadth. This \textit{general~+~expert} joint training paradigm—unified under the natural language action interface—enables stable co-training across diverse scenarios and tasks, with each domain's physical knowledge mutually reinforcing the others. Asymmetric 3D RoPE positional encoding and multi-view concatenation training enable geometrically consistent synthesis across synchronized camera views without architectural modification.

Evaluated on four established benchmarks, \textsc{Qwen-RobotWorld} achieves competitive performance across cross-scenario and cross-task settings. It outperforms all open-source models on WorldModelBench (8.99, 3rd overall), attaining perfect physics adherence scores across Newton's laws, mass conservation, fluid dynamics, and gravity---on par with leading closed-source models---while achieving strong instruction following (2.33/3.0). It ranks \textbf{1st overall} on EWMBench (4.60), with substantially leading motion fidelity in HSD (\textbf{0.566}, +33\% over the runner-up) and top scene consistency (0.914). On DreamGen Bench, the model ranks \textbf{1st overall} (4.952) across three robotic embodiment subsets, excelling in object-level compositional generalization. On PBench, it outperforms all open-source models (0.804), with domain understanding placing 3rd overall (0.857) and motion smoothness ranking 2nd among open-source models (0.990). Qualitative results further showcase generalization across cross-task video editing---including human-to-robot transfer, where the model synthesizes realistic robot execution from a human demonstration video without robot-specific prompting---as well as autonomous driving scene synthesis and room-scale indoor navigation generation; additional zero-shot performance on RoboTwin-IF benchmark further support robust transfer under complex instructions.

Our contributions are summarized as follows:
\begin{itemize}[leftmargin=*,topsep=2pt,itemsep=1pt]
  \item \textbf{Framework.} We propose \textsc{Qwen-RobotWorld}, a language-conditioned video world model that treats natural language as a universal action interface to unify cross-scenario and cross-task embodied capabilities. By jointly training manipulation, driving, navigation, and human-to-robot transfer under a shared language interface, the model achieves complementary physical generalization that no single-domain model can match.
  \item \textbf{Data.} We propose an action-language mapping framework that standardizes 20+ robot embodiment types and 500+ action categories into a unified natural language interface, and construct approximately 8.6M high-quality, cross-scenario, cross-task embodied video-text pairs constituting the EWK dataset.
  \item \textbf{Training.} We propose a general~+~expert joint training paradigm that, under the unified natural language interface, equips the model with both broad world modeling capability and deep embodied domain expertise, enabling stable and scalable co-training across diverse scenarios and tasks.
  \item \textbf{Performance.} \textsc{Qwen-RobotWorld} achieves comprehensive improvements on cross-scenario and cross-task embodied evaluation metrics, ranking 1st overall on EWMBench and DreamGen Bench and outperforming all open-source models on WorldModelBench and PBench.
\end{itemize}

\begin{figure}
    \centering
    \includegraphics[width=\linewidth]{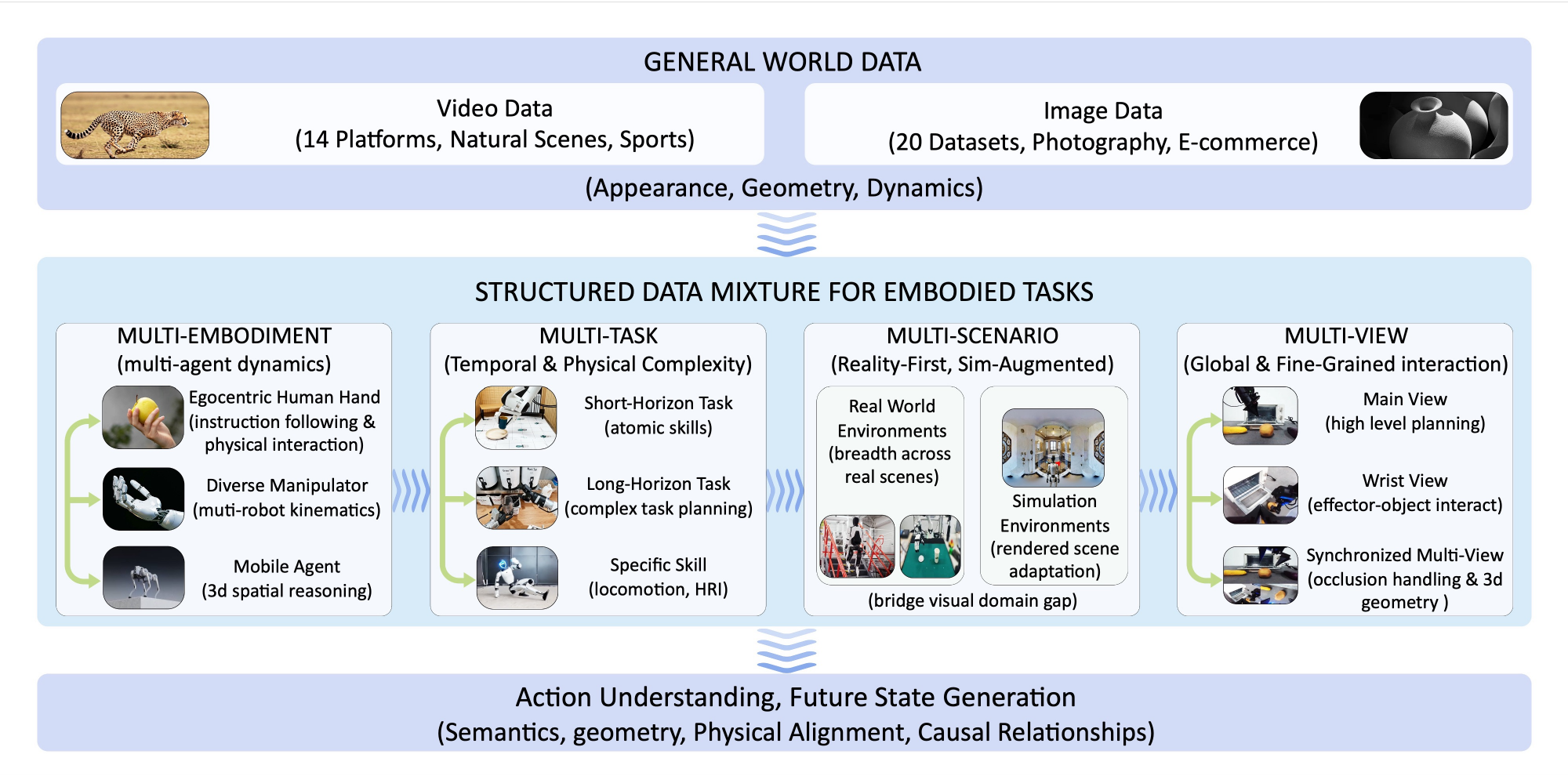}
    \vspace{-6mm}
    \caption{Overview of the Embodied World Knowledge (EWK) training corpus. \textbf{General world data} (top) supplies foundational priors on appearance, geometry, and dynamics from internet-scale video and image collections. \textbf{Structured embodied data} (middle) is organized along four complementary axes, each targeting a distinct source of physical variation: \textit{Multi-Embodiment} (human hands, diverse robot manipulators, mobile agents); \textit{Multi-Task} (short-horizon atomic skills, long-horizon compositional planning, specific skills such as locomotion and HRI); \textit{Multi-Scenario}, a \textit{reality-first, sim-augmented} design that bridges real captures and the simulators where downstream VLA policies are trained and evaluated; and \textit{Multi-View} (main, wrist, and synchronized multi-view streams covering both global planning and fine-grained effector--object interaction). Jointly, these signals supply the semantics, geometry, physical alignment, and causal relationships (bottom) required for language-conditioned action understanding and future-state generation.}
    \label{fig:dataset_visual}
    \vspace{-2mm}
\end{figure}

\section{Data}  \label{sec:data}

The central challenge in training a universal embodied world model is not data scale alone, but representational heterogeneity: robotic manipulation actions are expressed as joint angles or end-effector waypoints, driving as steering commands and velocity profiles, and navigation as heading vectors—each requiring a separate model or interface per domain. We resolve this through an \textbf{action-language mapping} framework that converts heterogeneous actions from 20+ robot embodiment types and 500+ action categories into a unified natural language interface. Under this unified interface, videos from a Franka gripper, an autonomous vehicle, and an indoor navigation agent all become instances of the same language-conditioned video generation task, enabling cross-scenario and cross-task joint training under a single model without any domain-specific control interface. As shown in Figure~\ref{fig:dataset_visual}, this framework produces approximately 6M high-quality, cross-scenario, cross-task embodied video–text pairs, which we further augment with general video data (30\% of the total) to construct the Embodied World Knowledge (EWK) dataset: a corpus of 8.6M video–text pairs comprising over 200M observation frames.

\subsection{Action-Language Mapping}
\label{sec:action_lang_mapping}

The action-language mapping framework addresses a fundamental asymmetry in embodied data: the \textit{visual} states (video frames) are already in a common pixel space, but the \textit{action} representations are fragmented across incompatible modalities. Our framework resolves this by projecting all action signals onto a shared natural language space, so that the same diffusion transformer can learn $s_{t+1} = f(s_t, a_t)$ regardless of the underlying physical domain.

\paragraph{Why Language as the Unified Action Interface.}
Unlike low-level action representations—joint angles, end-effector waypoints, force-torque commands—which are hardware-specific and require a separate control interface per embodiment, natural language offers a universal, embodiment-agnostic action interface. A single instruction such as ``grasp the red cup and lift it vertically'' implicitly encodes the full action sequence, goal state, and physical constraints, without any knowledge of the underlying kinematic chain. By training the model to predict the next visual state $s_{t+1}$ from a language action $a_t$ alone, we obtain a simulation backbone that generalizes across embodiments—whether a Franka gripper, an Aloha dual-arm system, or a humanoid—without retraining or re-engineering robot-specific interfaces. This generality, however, places demanding requirements on annotation quality: each caption must function as a \textit{complete, self-contained action specification}, precise enough that the model can predict $s_{t+1}$ from $a_t$ and $s_t$ alone, without access to any robot metadata or proprioceptive signals.

\paragraph{Hierarchical Five-Layer Annotation.}
To consistently produce such action-rich captions across 20+ robot embodiment types and 500+ action categories, we design a hierarchical annotation framework with five progressive layers. The first three form a structured chain-of-thought that decomposes each visual state transition into interpretable components:
\begin{enumerate}[leftmargin=*, topsep=2pt, itemsep=1pt]
    \item \textbf{Task Goal Layer}---infer the high-level intent of the transition (what should change between $s_t$ and $s_{t+1}$), integrating external instructions with observed video content;
    \item \textbf{Action Detail Layer}---decompose the action $a_t$ into spatio-temporal trajectories, micro-actions, speed, and force, with \textbf{mandatory explicit declaration of viewpoint information} (egocentric main view, wrist view, external view, or concatenated multi-view combinations);
    \item \textbf{Physical Feedback Layer}---describe the observable consequences of the action on the environment (object displacement, deformation, contact state changes), grounding each transition in verifiable physical outcomes.
\end{enumerate}
Based on this analysis, two granularities of action descriptions are generated:
\begin{enumerate}[leftmargin=*, topsep=2pt, itemsep=1pt]
    \setcounter{enumi}{3}
    \item \textbf{Comprehensive Description} (50--100 words)---fully specifies the viewpoint--agent--action--feedback quadruple, providing a rich action signal for precise state transition prediction;
    \item \textbf{Concise Description} (15--30 words)---retains only the essential viewpoint--agent--key action elements, enabling the model to handle brief, high-level commands at inference time.
\end{enumerate}
We enforce four quality control principles: \textit{operation focus} (only agent actions and object interactions), \textit{viewpoint definition} (explicit viewpoint type and semantic role), \textit{objectivity} (only visible dynamics), and \textit{physical verifiability} (only visually verifiable outcomes). In training, we sample from comprehensive and concise descriptions with equal probability (50\% each), so the model learns to execute both detailed trajectory specifications and brief task-level commands.

\paragraph{Coverage: 20+ Robot Embodiments, 500+ Action Categories.}
The framework is applied across all data domains. 
On the embodiment axis, it covers human hands, seven robot arm configurations (single-arm gripper, dual-arm gripper, single-arm dexterous hand, dual-arm dexterous hand, mobile dual-arm, half-humanoid, and full humanoid), ego vehicle (surround-view cameras), pedestrian/drone, and mobile navigation agent---representing 20+ distinct robot embodiments in total, as sourced from RoboCoin (15 robot models across three structural categories), Robomind (4 morphologies), InternData-A1 (4 robot models), Groot-XE, and various other datasets.
On the action axis, it spans 500+ action categories derived from the explicit motion primitive vocabularies across our training datasets---Agibot-World alone defines 84 distinct manipulation primitives (grasp, push, pour, fold, wipe, cut, etc.)---supplemented by unique primitives from other manipulation datasets and locomotion/navigation actions (turning, lane-changing, waypoint following, obstacle avoidance, etc.), organized into four tiers: (1) manipulation primitives, (2) long-horizon compositions, (3) locomotion and navigation, and (4) dynamic and deformable interactions.
This systematic coverage ensures that the resulting embodied video-text pairs span a semantically rich and physically diverse action space that no single domain could provide.

\subsection{Data Collection}

\begin{table*}[t]
\centering
\scriptsize
\caption{Detailed inventory of the Embodied World Knowledge (EWK) training data mixture, organized by domain.}
\vspace{-2mm}
\label{tab:data_inventory_detailed}
\resizebox{\textwidth}{!}{%
\begin{tabular}{@{} L{3.0cm} L{2.4cm} L{2.4cm} L{2.8cm} L{3.2cm} @{}}
\toprule
\textbf{Dataset} & \textbf{Embodiment} & \textbf{Views} & \textbf{Tasks} & \textbf{Contribution} \\
\midrule

% ========== A. MANIPULATION ==========
\rowcolor{blue!6} \multicolumn{5}{@{}l@{}}{\textbf{\textit{A. Manipulation}}} \\[2pt]

EgoHOD~\cite{pei2025egohod}, EPIC-Kitchens~\cite{damen2018epic}, Egocentric-10k~\cite{buildaiegocentric10k2025}
& Human hands
& Egocentric
& Daily grasping \& kitchen
& Dexterity \& coordination prior \\[3pt]

Bridge V2~\cite{walke2023bridgedata}, RH20T~\cite{fang2023rh20t}, Droid~\cite{khazatsky2024droid}
& Single-arm grippers
& external + wrist
& Tabletop pick-and-place
& Interaction primitives \\[3pt]

Robomind~\cite{wu2024robomind}, RoboCoin~\cite{wu2025robocoin}
& Single/dual-arm, humanoids
& Ego + external
& Rigid \& deformable objects
& Cross-embodiment generalization \\[3pt]

Agibot-World~\cite{bu2025agibotworld}, Galaxea~\cite{galaxea2025openworld}
& Single-arm (gripper + dexterous hand)
& Synced ego + wrist + external
& Long-horizon sequential
& Temporal \& multi-view consistency \\[3pt]

Qwen-Aloha (internal)
& Dual-arm grippers
& Head + dual wrist
& Diverse grasping
& Multi-view grasping prior \\[3pt]

ActionNet~\cite{fourier2025actionnet}, OpenLoong~\cite{openloong2025baihu}
& Dexterous hands
& Wrist + external
& Tool use \& in-hand
& Fine-grained dexterity \\[3pt]

InternData-A1~\cite{tian2025interndataa1}, Robotwin~\cite{chen2025robotwin}, Groot-XE~\cite{bjorck2025grootn1}, RT1~\cite{brohan2023rt1}
& Mixed arms (simulated)
& Variable
& Fluids \& deformables
& Sim-to-real alignment \\

\midrule

% ========== B. AUTONOMOUS DRIVING ==========
\rowcolor{green!6} \multicolumn{5}{@{}l@{}}{\textbf{\textit{B. Autonomous Driving}}} \\[2pt]

Waymo E2E~\cite{waymo_e2e}, NVIDIA PhysicalAI-AD~\cite{nvidia_physicalai}
& Ego vehicle
& 5--8 surround-view
& Urban driving \& traffic
& Large-scale motion \& 3D geometry \\[3pt]

Bench2Drive~\cite{bench2drive}
& Ego vehicle (sim)
& 6 surround-view
& 9.8K traffic scenarios
& Sim diversity \& GT annotations \\[3pt]

Sekai~\cite{sekai}
& Pedestrian / drone
& Egocentric
& Urban walking
& Pedestrian-scale locomotion \\

\midrule

% ========== C. INDOOR NAVIGATION ==========
\rowcolor{orange!6} \multicolumn{5}{@{}l@{}}{\textbf{\textit{C. Indoor Navigation}}} \\[2pt]

VLNVerse~\cite{lin2025vlnverse}
& Mobile agent
& Egocentric
& 134 indoor scenes, lang-guided
& 3D reasoning \& lang-trajectory align \\

\midrule

% ========== D. HUMAN-TO-ROBOT TRANSFER ==========
\rowcolor{violet!6} \multicolumn{5}{@{}l@{}}{\textbf{\textit{D. Human-to-Robot Transfer}}} \\[2pt]

Paired H2R dataset
& Human $\rightarrow$ 14 robot arms
& Egocentric bimanual
& Cross-embodiment manipulation
& Video editing supervision \\

\bottomrule
\end{tabular}%
}
\vspace{-4mm}
\end{table*}

\subsubsection{General Data}
General world data lays the foundation for the model to grasp basic physical laws and form accurate visual representations. This category encompasses diverse videos and still images from the internet. Video data are standardized to 24 FPS and support multiple resolutions and aspect ratios (1:1, 2:3, 3:2, 3:4, 4:3, 9:16, 16:9, etc.). Image data integrates high-quality static photographs, serving as visual quality anchors that establish precise representations of object appearance, material texture, and spatial composition. All general data is annotated with natural language descriptions generated by Qwen2.5-VL~\cite{Qwen2.5-VL}; annotations omit viewpoint-specific information to maintain flexibility and generality. Notably, we adopt a conservative stance on AI-generated content (AIGC): general data excludes AI-produced images and videos, as these often introduce visual artifacts, physical inconsistencies, and implicit biases that could undermine the model's generalization capabilities.

\subsubsection{Embodied Manipulation Data}

To enable the world model to acquire grounded physical understanding across scenarios and tasks, we build a structured data mixture spanning manipulation, driving, navigation, and cross-embodiment transfer domains, as summarized in Table~\ref{tab:data_inventory_detailed}. For the core manipulation domain, we organize the data around four dimensions: Multi-Embodiment, Multi-Task, Multi-Scenario, and Multi-View.

\paragraph{Multi-Embodiment.}
Manipulation data spans a spectrum of embodiments---human hands, single-arm grippers, dual-arm dexterous systems, mobile manipulators, and full-body humanoids---so the model learns to separate task-level intent from embodiment-specific kinematics. Human manipulation data (EgoHOD~\cite{pei2025egohod}, EPIC-Kitchens~\cite{damen2018epic}) provides a dexterity ceiling: the model observes what physically capable interaction looks like, acquiring priors for fluid hand--eye coordination and tool use. Robot data then teaches the model how those same intents map onto diverse mechanical morphologies. By exposing the model to both human demonstrations and robot executions of overlapping tasks (e.g., Robomind~\cite{wu2024robomind}, RoboCoin~\cite{wu2025robocoin}), it learns embodiment-invariant action semantics---the ability to predict ``what should happen next'' regardless of whether the actor is a two-finger gripper or a seven-finger dexterous hand.

\paragraph{Multi-Task.}
The manipulation corpus covers a skill hierarchy from atomic contact-level actions to extended multi-step procedures, teaching the model to operate at multiple temporal granularities. Short-horizon datasets (Bridge V2~\cite{walke2023bridgedata}, RH20T~\cite{fang2023rh20t}) provide dense coverage of fundamental interaction primitives---grasping, pushing, inserting---that ground the model's understanding of contact physics and object affordances. Long-horizon datasets (Agibot-World~\cite{bu2025agibotworld}, Galaxea~\cite{galaxea2025openworld}) chain these primitives into coherent sequences, forcing the model to maintain state tracking and causal reasoning across dozens of steps. Additionally, dynamic-interaction datasets (Humanoid Everyday~\cite{zhao2025humanoideveryday}) introduce high-velocity, whole-body motions that test the model's ability to predict outcomes under significant momentum and balance constraints. Together, this range ensures the model can reason about both ``what happens when you press here'' and ``what happens after ten sequential decisions.''

\paragraph{Multi-Scenario.}
Multi-scenario coverage advances along two complementary axes: \textit{breadth across real environments}, and \textit{extension to simulator-rendered scenarios}. Along the first axis, physical interaction manifests differently depending on context---a kitchen counter presents different lighting, clutter density, and surface properties than a factory floor or an outdoor worksite. Our manipulation data is therefore predominantly real-world, spanning domestic kitchens, workshops, laboratories, and unstructured outdoor settings, exposing the model to genuine variation in illumination, occlusion, material appearance, and background complexity---so it does not brittly overfit to any single environment. Along the second axis, we incorporate photorealistic simulation data (InternData-A1~\cite{tian2025interndataa1}) as a first-class complement. This is motivated by the VLA landscape: a substantial portion of policy models are trained in simulators, and virtually all are evaluated there using standardized benchmarks such as LIBERO~\cite{liu2024libero}, SimplerEnv~\cite{li2024simplerenv}, and RLBench~\cite{james2020rlbench}. A world model intended as a general simulation backbone must therefore generate faithfully under simulator-style appearances and physics, bridging the visual domain gap between real and synthetic observations so it can serve both sim-to-real transfer and closed-loop evaluation pipelines. The simulation portion additionally supplies precisely controlled variations in lighting, object pose, and camera placement that further strengthen visual robustness.

\paragraph{Multi-View.}
Single-view data teaches the model to predict plausible futures from a fixed perspective, but many physically critical events are partially or fully occluded from any single camera. Synchronized multi-view recordings (Agibot-World~\cite{bu2025agibotworld}, Robomind~\cite{wu2024robomind}) expose the model to the same event from head-mounted, wrist-mounted, and external viewpoints simultaneously. This serves two purposes: during training, cross-view correspondence acts as a geometric regularizer, implicitly teaching the model about object shape, depth, and spatial relationships; at inference, the model can generate from any individual viewpoint or compose multi-view outputs that remain mutually consistent. Approximately 1.6M of our 6M embodied samples include synchronized 2--4 view concatenations, providing substantial multi-view supervision without dominating the corpus.

\subsubsection{Autonomous Driving Data}
\label{sec:driving_data}

While manipulation data captures fine-grained object interactions within a confined workspace, autonomous driving data exposes the model to a substantially larger motion space with diverse maneuvers (turning, lane changing, acceleration) spanning a much wider range of velocities and trajectories. Driving scenes also contain rich multi-agent dynamics—surrounding vehicles, pedestrians, and cyclists interacting under traffic rules—requiring the world model to learn how multiple objects move, occlude, and influence each other over time. Furthermore, the large camera displacement provides dense supervisory signal for 3D scene geometry through parallax and perspective changes, strengthening the model's capacity for view synthesis and spatial reasoning.

We curate multi-view driving videos from four large-scale datasets: \textbf{Waymo E2E}~\cite{waymo_e2e} (real-world driving, 8 surround-view cameras, 7{,}044 clips / 11.3h), \textbf{NVIDIA PhysicalAI-AD}~\cite{nvidia_physicalai} (real-world driving, 5 cameras with 30$^\circ$--120$^\circ$ FoV, 1{,}342{,}418 clips / 1{,}715.9h), \textbf{Bench2Drive}~\cite{bench2drive} (CARLA-simulated driving under 9{,}881 diverse traffic scenarios, 6 cameras, 384{,}948 clips / 511.2h), and \textbf{Sekai}~\cite{sekai} (egocentric pedestrian walking and drone videos, 9{,}995 clips / 166.6h with scene and weather annotations). In total, the driving data comprises 1{,}744{,}405 clips spanning 2{,}405 hours. We apply a unified three-stage processing pipeline: (1)~frame extraction with trajectory unification into a common waypoint format, (2)~action-based clip segmentation (2--8\,s) according to ego maneuver transitions, and (3)~caption generation combining structured trajectory descriptions with optional VLM augmentation.

\subsubsection{Egocentric Indoor Navigation Data}
\label{sec:indoor_nav_data}

Egocentric indoor navigation data provides a complementary perspective to both manipulation and driving data. Unlike manipulation which focuses on fine-grained object interactions within a confined workspace, and driving which operates in large-scale outdoor environments, indoor navigation requires the model to understand room-scale spatial layouts, obstacle-aware path planning, and the mapping from textual navigation commands to spatially coherent visual trajectories.

Following VLNVerse~\cite{lin2025vlnverse}, we collect physically grounded egocentric navigation data using NVIDIA Isaac Sim~\cite{nvidia2022isaacsim} with photorealistic rendering and continuous control. We gather 6{,}064 successful navigation episodes across 134 indoor scenes, each consisting of an egocentric RGB video ($256 \times 256$ resolution at 10\,FPS) paired with natural language navigation instructions. The trajectories average approximately 8.2\,m in length (ranging from 4 to 17.5\,m), accumulating a total traversal distance of roughly 49.8\,km and approximately 5.8 hours of continuous first-person navigation video. The instructions are provided in two formats: single-string step-by-step directives (3{,}031 episodes, averaging 67.2 words) and multi-granularity descriptions at formal, natural, and casual registers (3{,}033 episodes).

Video generation models trained on such traversal data can acquire emergent 3D consistency and spatial coherence across frames~\cite{gao2026advancing,bar2025navigation}, while the physically grounded, action-conditioned nature of each sequence encourages the model to internalize depth reasoning, geometric consistency, and obstacle-aware planning~\cite{shang2025roboscape,han2025roomtour3d,zhen2025tesseract}. By grounding language instructions in continuous egocentric traversals, this data enables the world model to jointly learn language understanding, 3D spatial reasoning, and embodied action prediction within indoor environments.

\subsubsection{Human-to-Robot Transfer Data}
\label{sec:h2r_data}

To train the model on cross-embodiment visual correspondence without physical robot collection, we curate two complementary sources of human-to-robot transfer data. The first is a large-scale human-robot paired dataset constructed from egocentric bimanual manipulation recordings via an automated pipeline: 3D hand keypoints are extracted through MANO~\cite{romero2017mano} reconstruction and retargeted to robot end-effector trajectories, human hands are removed via video inpainting, and 14 robot arm models are rendered into the inpainted scene using MuJoCo~\cite{todorov2012mujoco} inverse kinematics, yielding four aligned video streams per episode (original human video, hand-removed scene, pure simulation, and robot-overlaid scene). The diversity of 14 embodiments within shared scenes ensures the editing capability generalizes across robot morphologies.

The second source addresses a fundamental limitation of direct rendering: simplified renderers ignore scene illumination, cast shadows, and material-dependent specular reflections, creating a photometric gap between rendered and real observations. To bridge this, we build upon the open-sourced InternA1 dataset~\cite{tian2025interndataa1}, which uses NVIDIA Isaac Sim~\cite{nvidia2022isaacsim} to provide photorealistic RGB observations with environment lighting and accurate shadows. Using the same dynamics parameters and robot URDFs, we render matched egocentric views in MuJoCo~\cite{todorov2012mujoco}—without lighting or shadow effects—producing paired samples that share identical geometry and viewpoint while differing in photometric realism. This paired data enables the model to learn the visual mapping between simplified rendering and photorealistic observations, covering Franka Emika Panda, AgileX Split Aloha, ARX Lift2, and AgiBot Genie1 across single-arm, dual-arm, mobile dual-arm, and humanoid configurations, with approximately 80K episodes spanning pick-and-place, articulated object manipulation, and multi-object rearrangement tasks.

\subsection{Data Processing}
\label{sec:data_processing}

\begin{figure*}[t]
    \centering
    \includegraphics[width=\linewidth]{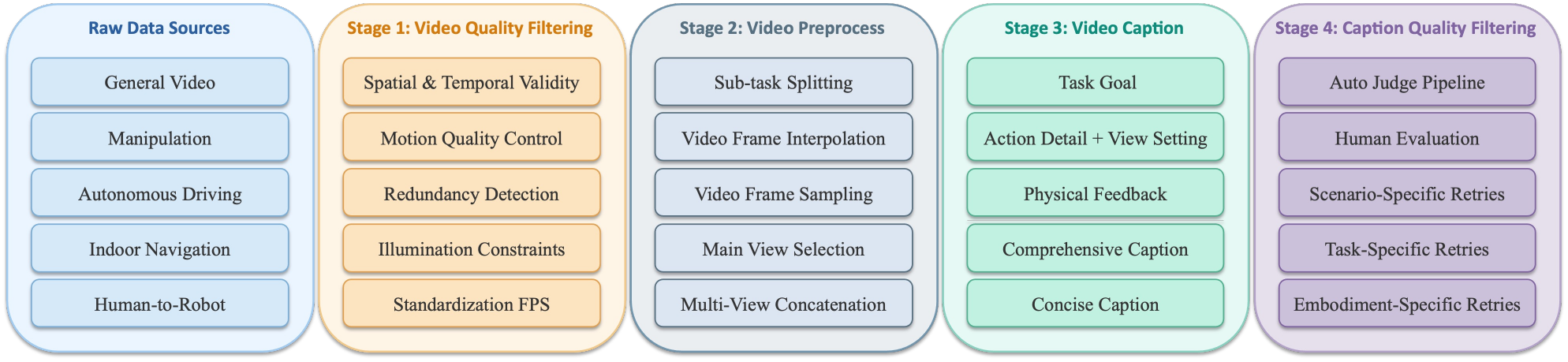}
    \caption{Overview of the unified data processing pipeline. Stage~1 (Raw Data Collection) collects heterogeneous data from five source categories spanning general and embodied domains. Stage~2 (Video Preprocessing) applies domain-adaptive operations---frame extraction, frame interpolation, sub-task splitting, main-view selection, and multi-view concatenation---to produce uniformly structured clips. Stage~3 (Hierarchical Annotation) generates viewpoint-aware captions through a five-layer framework: task goal, action detail, physical feedback, comprehensive caption, and concise caption. Stage~4 (Caption Quality Filtering) combines an automated LLM-based judge with human evaluation; underperforming captions are routed back for scenario-, task-, or embodiment-specific iterative prompt refinement.}
    \label{fig:data_pipeline}
    \vspace{-4mm}
\end{figure*}

We design a unified data processing pipeline that transforms heterogeneous raw data from diverse embodied and general video sources into high-quality, consistently formatted training samples. As illustrated in Figure~\ref{fig:data_pipeline}, the pipeline consists of four stages: \textbf{(1)}~Raw Data Collection, \textbf{(2)}~Video Preprocessing, \textbf{(3)}~Hierarchical Annotation, and \textbf{(4)}~Caption Quality Filtering with Iterative Prompt Refinement. Stages~2 and~3 apply domain-adaptive operations depending on source data characteristics, while Stage~4 forms a closed feedback loop that routes underperforming captions back for targeted re-annotation.

\subsubsection{Stage 1: Raw Data Collection}

The pipeline begins by ingesting raw video data from five source categories spanning both general and embodied domains. \textbf{General Video} provides internet-scale visual diversity from documentaries, professional stock libraries, and curated web clips. \textbf{Manipulation} data covers a broad spectrum of robot embodiments---single-arm grippers, dual-arm systems, dexterous hands, mobile platforms, and humanoids---from datasets including EgoHOD, Bridge V2, DROID, RoboMind, Agibot-World, and others. \textbf{Autonomous Driving} contributes large-scale ego-motion and multi-agent dynamics from Waymo, Bench2Drive, NVIDIA PhysicalAI-AD, and Sekai. \textbf{Indoor Navigation} supplies language-guided spatial reasoning episodes from VLNVerse across 134 indoor scenes. \textbf{Human-to-Robot Transfer} provides paired human demonstration and robot execution data constructed via our automated MANO-to-robot pipeline across 14 robot types.

\subsubsection{Stage 2: Video Preprocessing}

Raw videos undergo domain-adaptive preprocessing to produce uniformly structured clips suitable for training. We apply five complementary operations depending on the source data characteristics:

\textbf{Frame Extraction.}
For short-horizon task videos (typically single-step manipulations lasting 2--8\,s), we extract frames at a target rate that captures the essential phases of the interaction---approach, contact, manipulation, and result---ensuring each sample contains the complete causal chain of the atomic action.

\textbf{Frame Interpolation.}
When source videos have insufficient frame rates for smooth motion learning, we apply temporal interpolation to increase frame density, preserving continuous motion trajectories critical for modeling fine-grained contact dynamics and object-state transitions.

\textbf{Sub-task Splitting.}
For long-horizon episodes involving multi-step procedures (e.g., sequential pick-and-place, complex assembly), we decompose the video into semantically coherent sub-task segments. Each segment captures a complete atomic action with clear start and end states, preventing the partial-execution artifacts that arise from naive uniform truncation.

\textbf{Main-View Selection.}
For multi-camera recordings where only a primary viewpoint is needed (e.g., single-view manipulation training), we select the most informative camera stream---typically the egocentric or external view that best captures the interaction region---discarding redundant angles.

\textbf{Multi-View Concatenation.}
Conversely, for multi-view co-training, we concatenate synchronized clips from 2--4 camera viewpoints into a single horizontal layout, preserving temporal alignment across views. This enables the model to learn cross-view geometric consistency and synchronized state transitions without architectural modifications.

\subsubsection{Stage 3: Hierarchical Annotation}
\begin{AcademicBox}[Hierarchical Annotation Prompt Template]
\small
You are an expert embodied-AI annotator. Given a video clip of a \texttt{\{\{embodiment\_type\}\}} performing a manipulation task captured from a \texttt{\{\{viewpoint\_type\}\}} viewpoint, produce the following five-layer annotation.\\[4pt]
\textbf{--- Analysis Phase ---}\\[2pt]
\textbf{Layer 1 – Task Goal:} Identify the high-level intent of this interaction. What is the agent trying to achieve? Describe the desired state transition from the current observation to the goal state in one sentence.\\[2pt]
\textbf{Layer 2 – Action Detail:} Decompose the agent's actions into a step-by-step sequence. For each step, specify: (a) the motion trajectory and direction, (b) micro-actions (approach, grasp, lift, rotate, release, etc.), (c) estimated speed and force level. \textbf{You must explicitly state the viewpoint} (egocentric / wrist / external / multi-view concatenation).\\[2pt]
\textbf{Layer 3 – Physical Feedback:} Describe the observable physical consequences of each action on the environment: object displacement, deformation, contact state changes, and any secondary effects (e.g., liquid sloshing, cloth folding). Only include visually verifiable outcomes.\\[4pt]
\textbf{--- Generation Phase ---}\\[2pt]
\textbf{Layer 4 – Comprehensive Caption (50--100 words):} Synthesize Layers 1--3 into a cohesive paragraph that fully specifies the \textit{viewpoint--agent--action--feedback} quadruple. Include the camera perspective, embodiment identity, complete action sequence, and physical outcomes.\\[2pt]
\textbf{Layer 5 – Concise Caption (15--30 words):} Condense to an instruction-style summary retaining only the essential \textit{viewpoint--agent--key action} elements, suitable as a direct language command for the world model.\\[4pt]
\textbf{Quality Constraints:}\\
\textbullet~\textit{Operation focus}: describe only agent actions and object interactions; omit background narration.\\
\textbullet~\textit{Viewpoint definition}: explicitly name the viewpoint type and its semantic role.\\
\textbullet~\textit{Objectivity}: report only visible dynamics; do not infer hidden states.\\
\textbullet~\textit{Physical verifiability}: every claimed outcome must be visually confirmable from the video.
\end{AcademicBox}

Preprocessed videos pass through our five-layer hierarchical annotation framework (Section~\ref{sec:action_lang_mapping}), which generates viewpoint-aware captions at two granularities---comprehensive (50--100 words) and concise (15--30 words)---sampled with equal probability during training. The prompt template used by the annotation model is shown above.

\subsubsection{Stage 4: Caption Quality Filtering}

To ensure annotation quality across the diverse range of scenarios, tasks, and embodiments in our corpus, we implement a closed-loop quality filtering system combining automated assessment with human oversight. Captions that fail quality checks are routed back to Stage~3 for targeted re-annotation, forming an iterative refinement loop.

\textbf{Judge Pipeline.}
An automated LLM-based judge assesses each caption along several dimensions, including factual accuracy, specificity, instruction clarity, and viewpoint consistency. Specifically, it evaluates whether the caption correctly describes the video content, provides sufficient detail beyond generic descriptions, can function as an actionable command, and maintains spatial references consistent with the camera perspective. Captions that do not satisfy any of these criteria are flagged for further review.

\textbf{Human Evaluation.}
A subset of captions---particularly those near judgment thresholds or from underrepresented domains---undergoes manual review by human annotators who validate correctness, identify systematic failure patterns, and provide ground-truth corrections that inform subsequent prompt refinements.

\textbf{Iterative Prompt Refinement.}
When the judge pipeline identifies consistent underperformance in specific categories, we trigger targeted prompt redesign along three axes: \textbf{scenario-specific} retries (e.g., outdoor lighting conditions, kitchen environments), \textbf{task-specific} retries (e.g., articulated object manipulation, fluid pouring), and \textbf{embodiment-specific} retries (e.g., humanoid bimanual coordination, dexterous hand manipulation). Each retry employs a specialized prompt template tailored to the failure mode, and the refined captions are re-evaluated through the judge pipeline until they meet quality standards. This iterative loop ensures that no scenario, task, or embodiment category suffers from systematically poor annotations due to one-size-fits-all prompting.

\textbf{Final Corpus Statistics.}
After the complete four-stage pipeline, the final training corpus comprises approximately 8.6M video-text pairs (over 200M observation frames), with embodied data accounting for 70\% and general data for 30\%. Within the embodied portion, single-view manipulation data constitutes the majority at $\sim$4.3M samples, followed by $\sim$1.6M multi-view concatenated samples with synchronized 2--4 camera views, and $\sim$200K navigation and driving samples.

\section{Model}
\label{sec:model}

\subsection{Model Architecture}
\label{sec:model_arch}

\begin{figure}[h]
\centering
\includegraphics[width=1\linewidth]{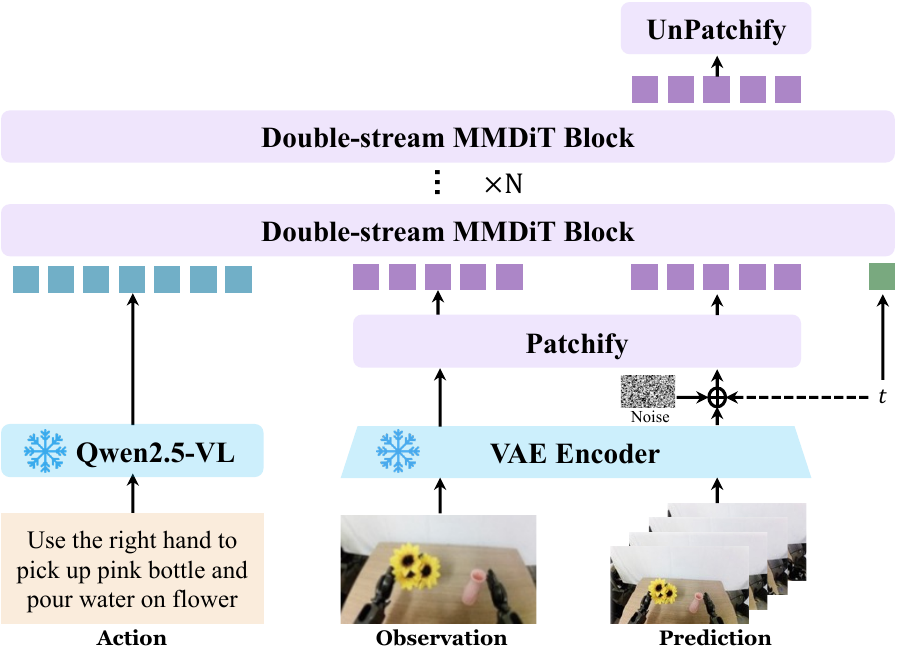}
   \caption{Overview of our video generation architecture with 60-layer double-stream MMDiT backbone.}
\label{fig:arc}
\end{figure}

As shown in Figure~\ref{fig:arc}, the model consists of three components: an MLLM as the action encoder, a VAE as the state encoder/decoder, and an MMDiT~\cite{esser2024scaling} as the transition function, organized in a double-stream design.

\textbf{MLLM — Action Encoder.} We employ a frozen Qwen2.5-VL~\cite{Qwen2.5-VL} to encode user inputs into condition signals. For a given input text $S$, it extracts last-layer hidden states $\mathbf{h} = \phi(S)$, serving as the action condition.

\textbf{VAE — State Encoder/Decoder.} The VAE encodes video frames into latent representations $\mathbf{z} = \mathcal{E}(\mathbf{x})$ and decodes predicted latents back into visual observations. We adopt the Wan-VAE~\cite{wan2025wan} architecture, which handles both image and video modalities.

\textbf{MMDiT — Transition Function.} The MMDiT adopts a double-stream architecture: the understanding stream receives the MLLM encoding $\mathbf{h}$ (projected via a trainable connector), and the generation stream receives noisy state latents from the VAE. At each block, the two streams interact via joint attention. The backbone comprises 60 double-stream blocks with 24 attention heads (head dimension 128), hidden size 3,072, and patch size 2×2. Total parameters: MLLM 7B, VAE 127M (encoder 54M + decoder 73M), MMDiT 20B. The context length supports up to 48,360 video tokens.

\subsection{3D Rotary Position Encoding}
\label{subsec:3d_rope}

We employ 3D RoPE~\cite{su2024roformer,heo2024rotary} to independently encode the temporal, spatial height, and spatial width dimensions. Rather than allocating dimensions uniformly, we use an asymmetric split: 16 dimensions for the temporal axis and 56 dimensions each for height and width, totaling 128 dimensions (\texttt{pe\_axes\_dim} = [16, 56, 56]). The temporal axis receives fewer dimensions as adjacent frames are strongly correlated; the spatial axes receive more to capture the greater diversity of object positions and scene layouts. We also apply Scalable RoPE~\cite{wan2025wan} to support generalization to varying resolutions and durations at inference.

\subsection{Scene2Robot}
\label{subsec:scene2robot}

\begin{figure}[t]
  \centering
  \includegraphics[width=\linewidth]{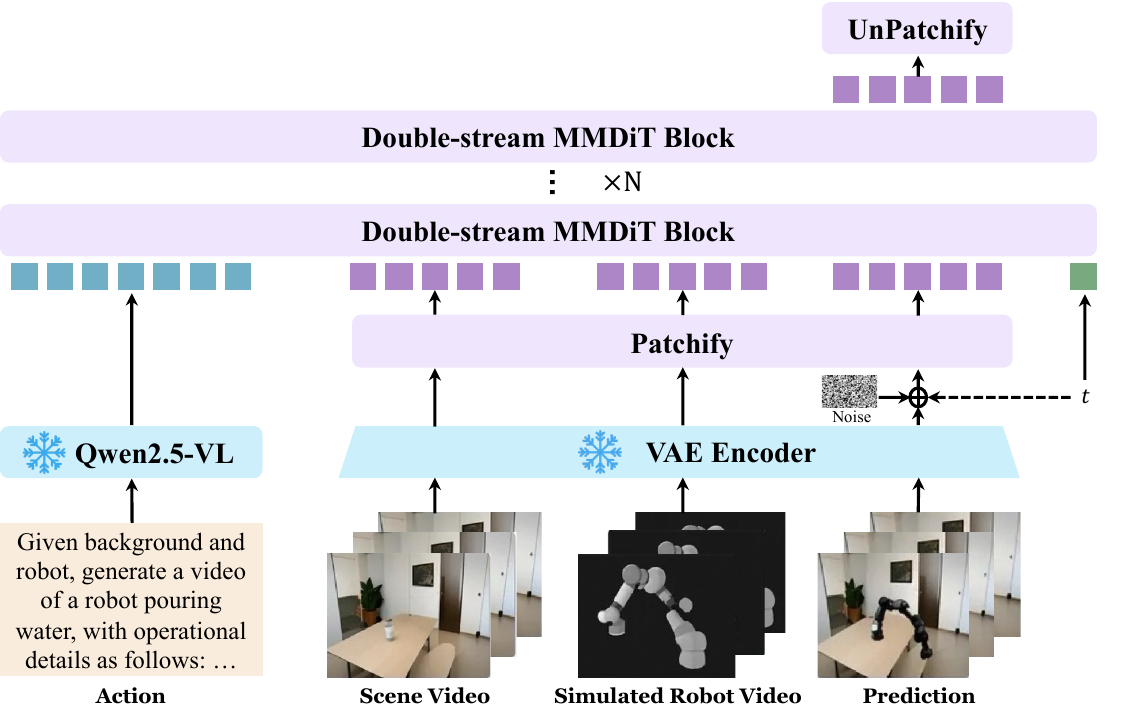}
  \vspace{-2mm}
  \caption{\textbf{Scene2Robot: multi-segment conditioning for cross-embodiment video synthesis.} The input sequence is organized as three contiguous segments — scene condition ($F$ frames), robot reference ($F$ frames), and generation ($F$ frames). An index-based mechanism assigns condition tokens to timestep $t=0$ and excludes them from loss computation, so only the generation segment is trainable. Joint attention at every MMDiT block enables the generation segment to simultaneously attend to scene appearance and robot motion trajectory, producing semantically coherent cross-embodiment synthesis.}
  \label{fig:scene2robot_architecture}
  \vspace{-4mm}
\end{figure}

Building upon the double-stream MMDiT architecture (\S\ref{sec:model_arch}) and the asymmetric 3D RoPE encoding (\S\ref{subsec:3d_rope}), we design \textsc{Scene2Robot}, a multi-segment conditioning mechanism that repurposes the same backbone for cross-embodiment video synthesis, as illustrated in Figure~\ref{fig:scene2robot_architecture}.

\textbf{First-Frame Conditioning (TI2V Baseline).}
For standard text-image-to-video tasks, the first frame serves as a fixed visual condition: its VAE latents are assigned timestep $t{=}0$ in the generation stream and excluded from the denoising loss, while the frozen Qwen2.5-VL encodes the text instruction into the understanding stream. Because the double-stream joint attention (\S\ref{sec:model_arch}) fuses both signals at every layer, the generation tokens can simultaneously attend to the visual anchor and the semantic action specification, producing temporally coherent continuations grounded in the language command.

\textbf{Multi-Segment Extension for Human-to-Robot Transfer.}
Human-to-robot transfer poses a video editing problem: the model must reference \textit{both} the scene context (background, object layout, lighting) \textit{and} the target robot's motion trajectory from a simulated demonstration. We address this by extending first-frame conditioning to a three-segment input sequence, all processed within the same VAE--MMDiT pipeline without any architectural modification:
\begin{enumerate}[leftmargin=*,topsep=2pt,itemsep=1pt]
    \item \textbf{Scene condition} ($F$ frames): the original human demonstration video, with human hands masked out, encoded by the VAE to provide appearance, spatial layout, and object state information.
    \item \textbf{Robot reference} ($F$ frames): a simulated robot execution rendered via MuJoCo, encoded by the VAE, supplying the target embodiment's kinematic trajectory and morphology.
    \item \textbf{Generation} ($F$ frames): noisy latents to be denoised into the final photorealistic robot execution video.
\end{enumerate}
Segments (1) and (2) share the same $t{=}0$ assignment as first-frame conditioning and are excluded from loss computation; only segment (3) receives gradient updates during training. The 3D RoPE encoding (\S\ref{subsec:3d_rope}) assigns each segment its own temporal index range, allowing the model to distinguish temporal positions across segments. Joint attention in every MMDiT block then enables the generation tokens to simultaneously attend to scene appearance from segment (1), robot motion from segment (2), and the MLLM action semantics from the understanding stream. This tripartite conditioning enables the model to synthesize photorealistic robot executions that faithfully preserve both the scene context and the instructed manipulation behavior.

\section{Training}
\label{sec:training}

\subsection{Training Strategy}

We propose a joint training paradigm in which general scene generation and robot manipulation prediction are unified under a single natural language interface as the same conditional video generation task, with the model continuously receiving gradient updates from both data regimes throughout training. This shared formulation allows general world priors and embodied action priors to reinforce each other through a common backbone, enabling stable cross-scenario and cross-task co-training. The curriculum proceeds in two progressive stages: pretraining establishes broad world foundations, and SFT deepens embodied specialization while preserving the general-expert balance.

\subsubsection{Pretraining Stage: Establishing General World Foundation}

\textbf{General World Priors.}
We curate over 200M real-world observation samples from 14 high-quality video platforms, covering natural scenes, daily life, and sports. This breadth allows the model to internalize domain-agnostic world priors—object motion, lighting variation, collision dynamics—that form the general backbone for later embodied generalization. We further incorporate multi-camera synchronized observations with 3D RoPE spatial encoding, establishing preliminary cross-view geometric consistency as a spatial foundation for multi-view embodied generation.

\textbf{Human Interaction Priors.}
We introduce large-scale first-person hand manipulation data (Ego4D~\cite{grauman2022ego4d}, EPIC-Kitchen~\cite{damen2018epic}, etc.). Human demonstration serves as a natural bridge between general and embodied: by learning grasping, tool use, and object manipulation from everyday human behavior, the model builds action priors and affordance understanding that transfer directly to robot operation in later stages.

\textbf{Multi-Task Joint Training.}
T2I, T2V, and TI2V tasks are trained jointly on a shared backbone, serving as the core mechanism through which general and embodied capabilities coexist in one model. The T2I task learns sharp visual representations from general image data, acting as a visual quality anchor whose object morphology knowledge automatically transfers to video generation tasks through the shared backbone, preventing deformation and identity inconsistency. Task ratios gradually shift from pure T2I toward full three-task joint training, so the model operates stably across multiple generation modes by the end of pretraining.

\subsubsection{SFT Stage: Embodied Specialization}

The SFT stage progressively deepens embodied expertise while keeping general world data in every training batch, ensuring that embodied specialization and general world modeling capability advance together rather than trade off.

\textbf{Progressive Embodied Knowledge Injection.}
We adopt a four-phase data mixing schedule. In early training, multi-embodiment robot data and human hand manipulation data co-dominate: human action priors guide the learning of cross-embodiment operation commonalities, while robot data strengthens concrete execution representations. We then gradually increase wrist-view and third-person view data to broaden viewpoint coverage. Building on this, we introduce multi-view concatenated training: synchronized first frames from multiple cameras are spatially concatenated as a single input, requiring the model to jointly generate subsequent frames for all views simultaneously, forcing the attention layers to establish cross-view spatial correspondences and achieve geometrically consistent multi-view generation. In the final phase, scarce high-complexity tasks (pouring, folding, bimanual coordination, multi-material interaction) and long-horizon reasoning data are targeted for supplementation to push the frontier of embodied capability. Throughout this process, general world data continuously participates in every training batch, jointly acting on the same backbone alongside embodied data to ensure that embodied specialization and general world modeling capability advance together.

\subsection{Training Objective and Infrastructure}

We adopt the flow matching objective~\cite{lipman2023flow,liu2022flow}, where input videos are encoded into latent space via the VAE encoder and noise is sampled from a standard normal distribution. Qwen2.5-VL encodes text inputs as guidance signal. Timesteps are sampled from a log-normal distribution with adaptive shifting based on video sequence length~\cite{esser2024scaling}. For TI2V tasks, the first-frame timestep is fixed at 0 to ensure that the generation process is conditioned on the given observation frame. Training is conducted with Megatron-LM~\cite{shoeybi2019megatron} using a hybrid parallelism strategy, with selective activation recomputation~\cite{korthikanti2023reducing} applied to a subset of dual-stream blocks to balance memory usage and training throughput.

\section{Experiments}
\label{sec:exp}

We conduct comprehensive evaluations on four benchmarks spanning embodied manipulation, physical reasoning, and general video quality. Across these benchmarks, our model delivers consistently strong results, achieving state-of-the-art performance on EWMBench for embodied world modeling (Overall \textbf{4.60}, +0.55 over LVP), ranking \textbf{1st overall} on DreamGen Bench (Total \textbf{4.952}), and \textbf{1st among open-source models} on WorldModelBench (Total \textbf{8.99}).

\noindent\textbf{Quantitative Evaluation} (\S\ref{sec:quantitative}).
We evaluate against two categories of baselines: (1) \textit{general video generation models}---Sora2~\cite{sora2}, Veo3~\cite{veo3}, Wan2.6~\cite{wan26}, Kling~\cite{kling}, and LTX-2~\cite{ltx2}; and (2) \textit{embodied world models}---Cosmos~\cite{cosmos}, WoW~\cite{wow}, LVP~\cite{lvp}, Vidar~\cite{vidar}, and GigaWorld~\cite{gigaworld}.

\noindent\textbf{Qualitative Analysis} (\S\ref{sec:qualitative}).
We evaluate manipulation capabilities along three progressive dimensions: fine-grained language grounding, generalization across embodiments, tasks, and viewpoints, and zero-shot robustness against strong baselines. 

\noindent\textbf{Cross-Domain Generalization} (\S\ref{sec:cross_domain}) further covers human-to-robot transfer, autonomous driving, and indoor navigation as supplementary tasks.

%=============================================================================
% QUANTITATIVE EVALUATION
%=============================================================================
\subsection{Quantitative Evaluation}
\label{sec:quantitative}

Unless noted otherwise, quantitative tables use \textbf{boldface} for the best value in each column and \underline{underline} for the second best.

%-----------------------------------------------------------------------------
\subsubsection{EWMBench: Embodied Motion Fidelity}
\label{sec:ewmbench}

\textbf{Benchmark.} EWMBench~\cite{yue2025ewmbench} evaluates embodied world models on three dimensions:
\textit{scene consistency} (SceneC),
\textit{motion correctness} (HSD, Dyn, nDTW),
and \textit{semantic alignment} (Diversity, BLEU, CLIP, Logics).
The benchmark contains 21 samples across 7 tasks with clear action-ordering constraints.

\begin{table}[h]
\centering
\caption{Performance comparison on EWMBench.}
\label{tab:ewmbench_results}
\vspace{-2mm}
\resizebox{\textwidth}{!}{
\begin{tabular}{ll|c|ccc|cccc|c}
\toprule
\multirow{2}{*}{\textbf{Type}} & \multirow{2}{*}{\textbf{Model}}
  & \textbf{Scene} & \multicolumn{3}{c|}{\textbf{Motion}} & \multicolumn{4}{c|}{\textbf{Semantics}} & \multirow{2}{*}{\textbf{Overall}} \\
\cmidrule(lr){3-3}\cmidrule(lr){4-6}\cmidrule(lr){7-10}
& & SceneC & HSD & Dyn & nDTW & Diversity & BLEU & CLIP & Logics & \\
\midrule
\multirow{5}{*}{General}
  & Veo3      & 0.8415 & 0.2130 & 0.1932 & 0.1613 & 0.0221 & 0.2139 & 0.8965 & 0.9474 & 3.49 \\
  & Wan2.6    & 0.6712 & 0.2034 & 0.0900 & 0.1715 & 0.0502 & 0.1616 & 0.8743 & \textbf{1.0000} & 3.22 \\
  & Kling26   & 0.8211 & 0.3272 & 0.1822 & 0.3423 & 0.0173 & \textbf{0.2591} & \underline{0.9014} & \textbf{1.0000} & 3.85 \\
  & LTX-2     & 0.7850 & 0.2076 & 0.1283 & 0.2443 & 0.0120 & 0.1425 & 0.8869 & 0.5000 & 3.01 \\
  & Sora2     & 0.8526 & 0.2807 & \textbf{0.3494} & 0.2754 & 0.0314 & \underline{0.2466} & \textbf{0.9100} & 0.9474 & 3.89 \\
\midrule
\multirow{5}{*}{Embodied}
  & Cosmos    & 0.7963 & 0.2500 & 0.2052 & 0.2533 & \textbf{0.0803} & 0.1230 & 0.8458 & 0.7333 & 3.29 \\
  & GigaWorld & 0.8707 & 0.3050 & 0.0849 & 0.2783 & 0.0278 & 0.2048 & 0.8873 & 0.9000 & 3.56 \\
  & LVP       & 0.8795 & \underline{0.4248} & 0.0433 & \underline{0.6226} & 0.0093 & 0.2179 & 0.8995 & \underline{0.9524} & \underline{4.05} \\
  & Vidar     & 0.7341 & 0.1877 & 0.1520 & 0.1769 & \underline{0.0653} & 0.1607 & 0.8821 & 0.9411 & 3.30 \\
  & Wow       & \underline{0.8866} & 0.2494 & 0.0529 & 0.2566 & 0.0266 & 0.1932 & 0.9001 & \underline{0.9524} & 3.52 \\
\midrule
  & \textbf{Ours} & \textbf{0.9142} & \textbf{0.5660} & \underline{0.3429} & \textbf{0.6708} & 0.0114 & 0.2079 & 0.8834 & \textbf{1.0000} & \textbf{4.60} \\
\bottomrule
\end{tabular}
}
\vspace{-2mm}
\end{table}

\textbf{Results.} Table~\ref{tab:ewmbench_results} shows our model ranks \textbf{1st} overall with a score of \textbf{4.60}, outperforming the runner-up LVP (4.05) by $+0.55$.
We lead in motion fidelity---HSD (\textbf{0.566}) surpasses LVP (0.425) by 33\%---and achieve top performance in scene consistency (SceneC: \textbf{0.914}) and logic constraint satisfaction (Logics: \textbf{1.00}).

%-----------------------------------------------------------------------------
\subsubsection{DreamGen Bench}
\label{sec:dreamgen}

\textbf{Benchmark.} DreamGen Bench~\cite{dreamgen} evaluates the quality of robot videos generated by video world models, measuring instruction following (IF) and physics alignment (PA) across three subsets of the GR1 robot embodiment: environment generalization (GR1-Env), object generalization (GR1-Object), and behavior generalization (GR1-Behavior).
IF is assessed using Qwen2.5-VL~\cite{Qwen2.5-VL} as the evaluator.

\begin{table}[h]
\centering
\caption{Performance comparison on DreamGen Bench.}
\label{tab:dreamgen_results}
\vspace{-2mm}
\resizebox{0.65\textwidth}{!}{
\begin{tabular}{l|cc|cc|cc|c}
\toprule
\multirow{2}{*}{\textbf{Model}}
  & \multicolumn{2}{c|}{\textbf{GR1-Env}}
  & \multicolumn{2}{c|}{\textbf{GR1-Object}}
  & \multicolumn{2}{c|}{\textbf{GR1-Behavior}}
  & \multirow{2}{*}{\textbf{Total}} \\
\cmidrule(lr){2-3}\cmidrule(lr){4-5}\cmidrule(lr){6-7}
& PA & IF & PA & IF & PA & IF & \\
\midrule
  Cosmos-sft & 0.709 & 0.655 & \underline{0.775} & 0.720 & 0.649 & 0.621 & 4.129 \\
  LVP       & \underline{0.810} & 0.772 & 0.745 & 0.829 & 0.713 & \textbf{0.889} & \underline{4.758} \\
  Vidar     & 0.445 & 0.647 & 0.478 & 0.726 & 0.394 & 0.651 & 3.341 \\
  GigaWorld & 0.621 & \textbf{0.933} & 0.500 & \underline{0.852} & 0.426 & \underline{0.884} & 4.216 \\
  Wow       & 0.793 & \underline{0.826} & 0.755 & 0.849 & \textbf{0.809} & 0.696 & 4.728 \\
\midrule
  \textbf{Ours} & \textbf{0.828} & 0.793 & \textbf{0.840} & \textbf{0.878} & \underline{0.781} & 0.832 & \textbf{4.952} \\
\bottomrule
\end{tabular}
}
\vspace{-2mm}
\end{table}

\textbf{Results.} Table~\ref{tab:dreamgen_results} shows our model achieves the highest total score of \textbf{4.952}, ranking \textbf{1st} overall.
We lead in GR1-Object IF (\textbf{0.878}, 1st), demonstrating strong object-level compositional generalization, and physics alignment is consistent across all subsets (PA: 0.828/0.840/0.781).
GR1-Behavior IF (0.832) slightly trails LVP (\textbf{0.889}) and GigaWorld (\underline{0.884}), indicating long-horizon behavior generalization as a direction for further improvement.

%-----------------------------------------------------------------------------
\subsubsection{PBench: Physical Behavior Evaluation}
\label{sec:pbbench}

\textbf{Benchmark.} PBench~\cite{nvidia2025pbench} evaluates models on two complementary aspects: (1) \textit{Domain Score}, which measures physical behavior understanding via QA pairs assessed by Qwen2.5-VL across six domains (AV, Robot, Industry, Physics, Human, Common Sense); and (2) \textit{Quality Score}, which measures visual quality via eight VBench~\cite{huang2024vbench} metrics including image-to-video consistency, aesthetic quality, motion smoothness, and subject consistency. The \textit{Overall Score} is the average of the two.

\begin{table}[h]
\centering
\caption{Performance comparison on PBench.}
\label{tab:pbbench_results}
\vspace{-2mm}
\resizebox{\textwidth}{!}{
\begin{tabular}{ll|cccccccc|c|c|c}
\toprule
\multirow{2}{*}{\textbf{Type}} & \multirow{2}{*}{\textbf{Model}}
  & \multicolumn{8}{c|}{\textbf{Quality Metrics (VBench)}}
  & \multirow{2}{*}{\textbf{Qual.}}
  & \multirow{2}{*}{\textbf{Domain}}
  & \multirow{2}{*}{\textbf{Overall}} \\
\cmidrule(lr){3-10}
& & I2V-Bg & I2V-S & Aes & Img & Bg-Con & Mot & Sub-Con & O-Con & & & \\
\midrule
\multirow{5}{*}{General}
  & Veo3    & 0.975 & \underline{0.980} & \textbf{0.526} & 0.698 & \underline{0.938} & \textbf{0.994} & 0.927 & 0.128 & \underline{0.771} & \textbf{0.882} & \textbf{0.827} \\
  & Wan2.6  & 0.856 & 0.843 & 0.514 & \textbf{0.719} & 0.906 & 0.978 & 0.843 & \underline{0.136} & 0.724 & 0.832 & 0.778 \\
  & Sora2   & \underline{0.981} & 0.973 & 0.487 & 0.672 & \textbf{0.961} & \textbf{0.994} & \underline{0.954} & 0.129 & 0.769 & 0.841 & 0.805 \\
  & Kling26 & \textbf{0.982} & 0.979 & \underline{0.521} & \underline{0.699} & 0.920 & 0.990 & 0.927 & 0.124 & 0.768 & \underline{0.874} & \underline{0.821} \\
  & LTX-2   & 0.948 & 0.955 & 0.506 & 0.622 & 0.932 & 0.986 & 0.904 & 0.118 & 0.746 & 0.845 & 0.796 \\
\midrule
\multirow{5}{*}{Embodied}
  & Cosmos    & 0.974 & 0.973 & 0.470 & 0.663 & 0.940 & 0.989 & 0.931 & \textbf{0.160} & 0.763 & 0.840 & 0.802 \\
  & LVP       & 0.979 & \textbf{0.981} & 0.515 & 0.679 & 0.954 & \underline{0.991} & \textbf{0.962} & 0.116 & \textbf{0.772} & 0.812 & 0.792 \\
  & GigaWorld & 0.957 & 0.944 & 0.495 & 0.641 & 0.925 & 0.984 & 0.892 & 0.128 & 0.746 & 0.841 & 0.794 \\
  & Vidar     & 0.935 & 0.922 & 0.501 & 0.573 & 0.912 & 0.982 & 0.863 & 0.120 & 0.726 & 0.810 & 0.768 \\
  & Wow       & 0.967 & 0.957 & 0.517 & 0.689 & 0.941 & 0.980 & 0.929 & 0.111 & 0.761 & 0.786 & 0.774 \\
\midrule
  & \textbf{Ours} & 0.956 & 0.943 & 0.455 & 0.649 & \underline{0.956} & 0.990 & 0.933 & 0.124 & 0.751 & 0.857 & 0.804 \\
\bottomrule
\end{tabular}
}
\vspace{-2mm}
\end{table}

\textbf{Results.} As shown in Table~\ref{tab:pbbench_results}, our model outperforms all among open-source models with an overall score of \textbf{0.804}.
Domain understanding is our strongest dimension (0.857, \textbf{3rd overall}), surpassing most closed-source models.
Motion smoothness also stands out (0.990, \textbf{2nd among open-source models}), reflecting consistent temporal coherence in generation.
Aesthetic quality (0.455) and imaging quality (0.649) are relatively lower, primarily because our model is purpose-built for embodied tasks and operates at a lower output resolution than general-purpose video generators, which reduces VBench's pixel-level quality scores; nonetheless, this resolution is fully sufficient for downstream robot control tasks.

%-----------------------------------------------------------------------------
\subsubsection{WorldModelBench: Physical Reasoning and Instruction Following}
\label{sec:worldmodelbench}

\textbf{Benchmark.} WorldModelBench~\cite{worldmodelbench} evaluates models on three dimensions: \textit{instruction following} (0--3 scale), \textit{common sense} (frame and temporal quality), and \textit{physics adherence} (5 violation types: Newton's laws, mass conservation, fluid dynamics, penetration, gravity).
The benchmark contains 350 instances across 7 domains with 56 subdomains.

\begin{table}[h]
\centering
\caption{Performance comparison on WorldModelBench.}
\label{tab:worldmodelbench_results}
\vspace{-2mm}
\resizebox{\textwidth}{!}{
\begin{tabular}{ll|c|ccc|ccccc|c|c}
\toprule
\multirow{2}{*}{\textbf{Type}} & \multirow{2}{*}{\textbf{Model}}
  & \textbf{Instr.}
  & \multicolumn{3}{c|}{\textbf{Common Sense}}
  & \multicolumn{5}{c|}{\textbf{Physics Adherence}}
  & \textbf{Phys.}
  & \multirow{2}{*}{\textbf{Total}} \\
\cmidrule(lr){3-3}\cmidrule(lr){4-6}\cmidrule(lr){7-11}\cmidrule(lr){12-12}
& & (0-3) & Frame & Temp & Overall & Newton & Mass & Fluid & Penetr. & Grav. & Overall & \\
\midrule
\multirow{5}{*}{General}
  & Veo3    & \textbf{2.52} & \underline{0.98} & \underline{0.95} & 1.93 & \textbf{1.00} & 0.89 & 0.99 & 0.91 & \textbf{1.00} & 4.80 & \underline{9.25} \\
  & Wan2.6  & \underline{2.50} & 0.99 & \underline{0.95} & \underline{1.94} & \textbf{1.00} & 0.89 & 0.99 & \underline{0.94} & \textbf{1.00} & \underline{4.83} & \textbf{9.27} \\
  & Sora2   & 2.21 & 0.96 & 0.93 & 1.88 & \textbf{1.00} & 0.91 & 0.99 & 0.95 & \textbf{1.00} & 4.84 & 8.93 \\
  & Kling26 & 1.59 & 0.97 & \textbf{1.00} & \textbf{1.97} & \textbf{1.00} & \textbf{1.00} & \textbf{1.00} & \textbf{1.00} & \textbf{1.00} & \textbf{5.00} & 8.55 \\
  & LTX-2   & 1.97 & 0.69 & 0.62 & 1.32 & 0.99 & 0.60 & \textbf{1.00} & 0.73 & \textbf{1.00} & 4.32 & 7.61 \\
\midrule
\multirow{5}{*}{Embodied}
  & Cosmos    & 2.14 & \textbf{1.00} & 0.94 & \underline{1.94} & \textbf{1.00} & 0.92 & \textbf{1.00} & 0.94 & \textbf{1.00} & 4.86 & 8.94 \\
  & LVP       & 2.01 & 0.89 & 0.91 & 1.80 & \textbf{1.00} & \underline{0.93} & 0.99 & 0.95 & \textbf{1.00} & 4.87 & 8.67 \\
  & GigaWorld & 2.13 & 0.59 & 0.46 & 1.05 & \textbf{1.00} & 0.48 & 0.99 & 0.69 & 0.98 & 4.13 & 7.31 \\
  & Vidar     & 1.62 & 0.54 & 0.45 & 0.99 & \textbf{1.00} & 0.56 & \textbf{1.00} & 0.85 & \textbf{1.00} & 4.40 & 7.01 \\
  & Wow       & 2.05 & 0.76 & 0.65 & 1.41 & \textbf{1.00} & 0.65 & 0.99 & 0.81 & \textbf{1.00} & 4.45 & 7.91 \\
\midrule
  & \textbf{Ours} & 2.33 & 0.87 & 0.85 & 1.72 & \textbf{1.00} & \textbf{1.00} & \textbf{1.00} & 0.94 & \textbf{1.00} & \underline{4.94} & 8.99 \\
\bottomrule
\end{tabular}
}
\vspace{-2mm}
\end{table}

\textbf{Results.} Table~\ref{tab:worldmodelbench_results} shows our model outperforms all open-source models (\textbf{8.99}, 3rd overall), trailing only closed-source Wan2.6 and Veo3.
We achieve perfect physics adherence (1.00) across all four categories and strong instruction following (2.33/3.0), with the common-sense gap attributable to our lower output resolution.

%=============================================================================
% QUALITATIVE ANALYSIS
%=============================================================================
\begin{figure*}[t]
    \centering
    \includegraphics[width=\linewidth]{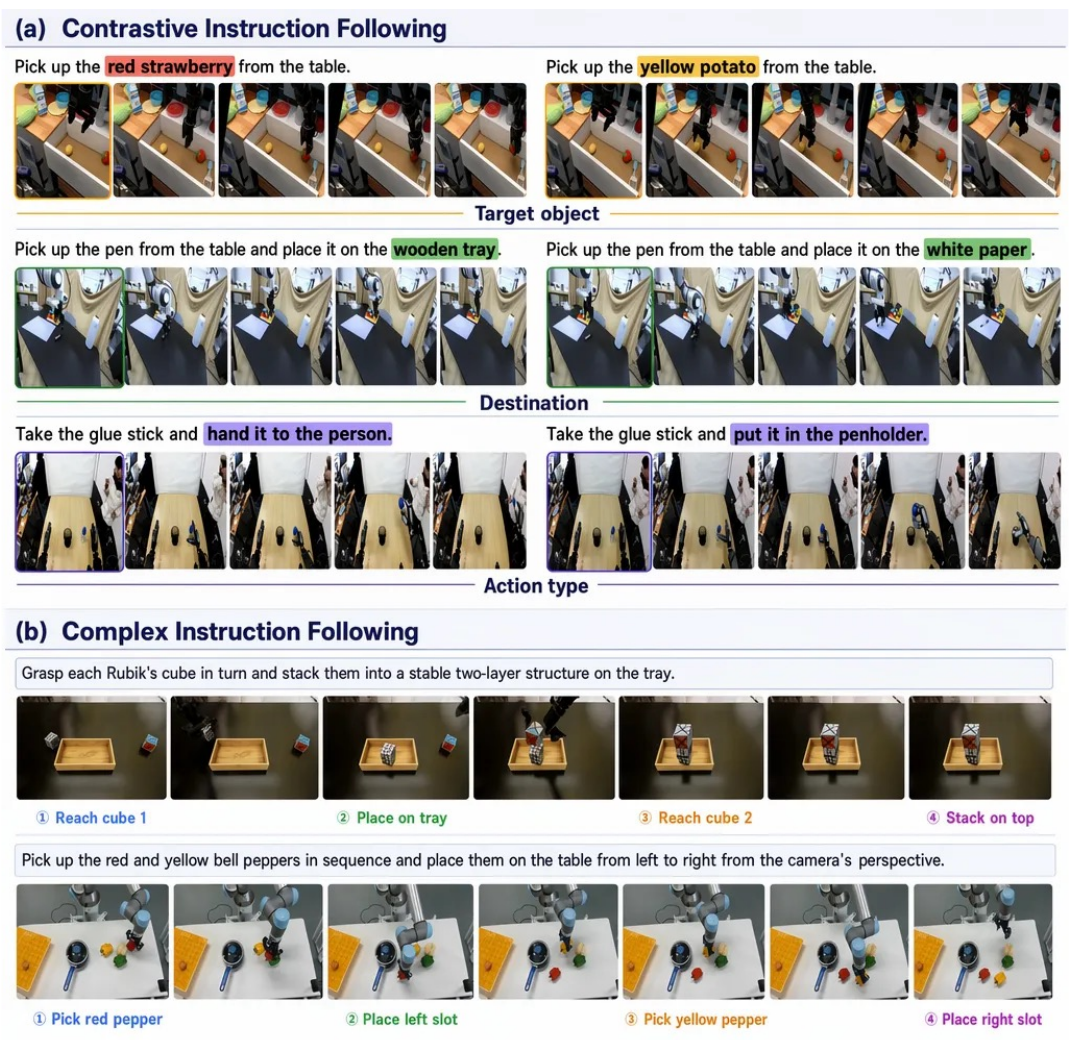}
    \vspace{-6mm}
    \caption{
        \textbf{Fine-grained language grounding.}
        \textbf{(a)~Contrastive:} each pair of columns shares an identical initial frame (colored border);
        only the highlighted keyword differs between the two instructions.
        \textit{Pair~1}: target object identity.
        \textit{Pair~2}: destination.
        \textit{Pair~3}: action type.
        In every case the generated motion is precisely grounded to the discriminating keyword.
        \textbf{(b)~Complex:} two examples requiring multi-step execution or abstract goal inference.
        Colored labels mark key action milestones within each generated sequence.
    }
    \label{fig:instruction_following}
    % \vspace{-4mm}
\end{figure*}

\subsection{Qualitative Analysis}
\label{sec:qualitative}

\subsubsection{Fine-Grained Language Grounding}

Precise grounding of language in visual actions is foundational to \textsc{Qwen-RobotWorld}'s design as a language-conditioned world model.
Figure~\ref{fig:instruction_following} evaluates this capability at two levels.
\textit{(a)~Contrastive pairs:} given identical initial frames, the model produces qualitatively distinct videos when a single keyword differs---target object identity, action type, or spatial placement---demonstrating fine-grained semantic discrimination beyond generic manipulation priors.
\textit{(b)~Complex instructions:} the model handles long-horizon sequential tasks with multi-step dependencies and abstract goal instructions that require inferring the manipulation sequence from context, decomposing each into a temporally coherent execution without explicit sub-task prompts.

\subsubsection{Generalization across Embodiments, Tasks, and Viewpoints}

\begin{figure}[t]
    \centering
    \includegraphics[width=\linewidth]{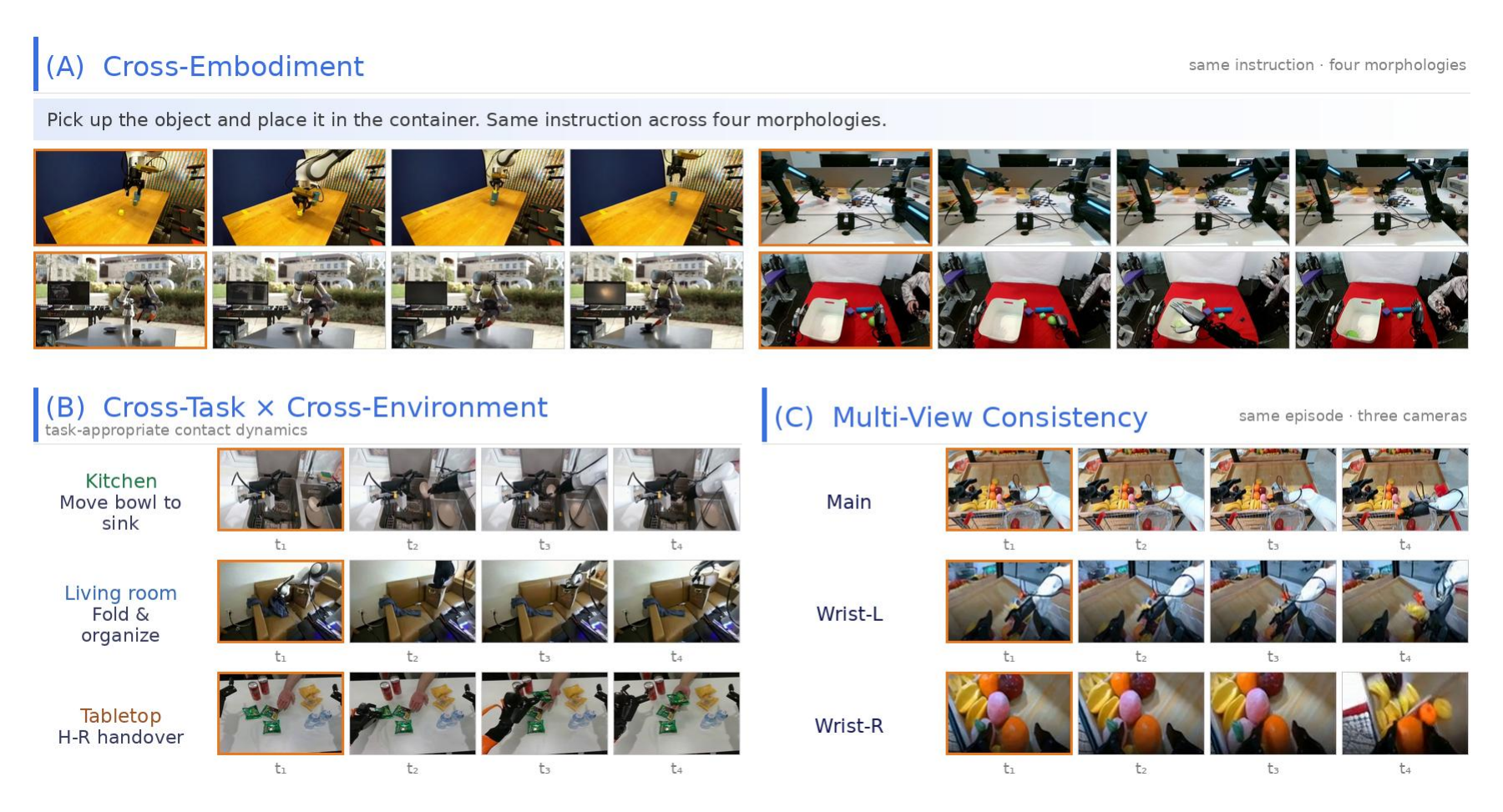}
    \vspace{-10mm}
    \caption{
        \textbf{Generalization across embodiments, tasks, and viewpoints.}
        \textbf{(A)~Cross-embodiment:} one instruction drives four morphologies (single-arm, dual-arm, humanoid, dexterous hand); three key frames per cell.
        \textbf{(B)~Cross-task $\times$ cross-environment:} initial frame (orange border) followed by four generated frames across four tasks.
        \textbf{(C)~Multi-view:} main and wrist cameras jointly generated from the same episode as (B, row\,1).
    }
    \label{fig:manipulation_gallery}
    \vspace{-4mm}
\end{figure}

Figure~\ref{fig:manipulation_gallery} demonstrates three complementary dimensions of \textsc{Qwen-RobotWorld}'s manipulation capability.
\textbf{(A)~Cross-embodiment:} a single instruction drives four distinct robot morphologies---single-arm gripper, dual-arm system, humanoid, and dexterous hand---without embodiment-specific adaptation, validating natural language as a universal action interface; each cell shows three key frames (initial, mid, final).
\textbf{(B)~Cross-task $\times$ cross-environment:} generations across fruit pick-and-place, bowl retrieval, cloth folding, and human--robot handover each exhibit task-appropriate contact dynamics, reflecting grounded physical knowledge across diverse real-world environments; each row shows an initial frame followed by four evenly-spaced generated frames.
\textbf{(C)~Multi-view consistency:} three synchronized camera streams (main, wrist-left, wrist-right) are jointly generated from the same supermarket pick-and-place episode as (B, row\,1), with object identity and motion trajectory remaining geometrically consistent across all viewpoints.

%=============================================================================
% RoboTwin-IF zero-shot robustness and diagnostics
%=============================================================================
\subsubsection{Zero-Shot Robustness on RoboTwin-IF}
\label{sec:lamv_qual}

\begin{figure}[H]
    \centering
    \includegraphics[width=\linewidth]{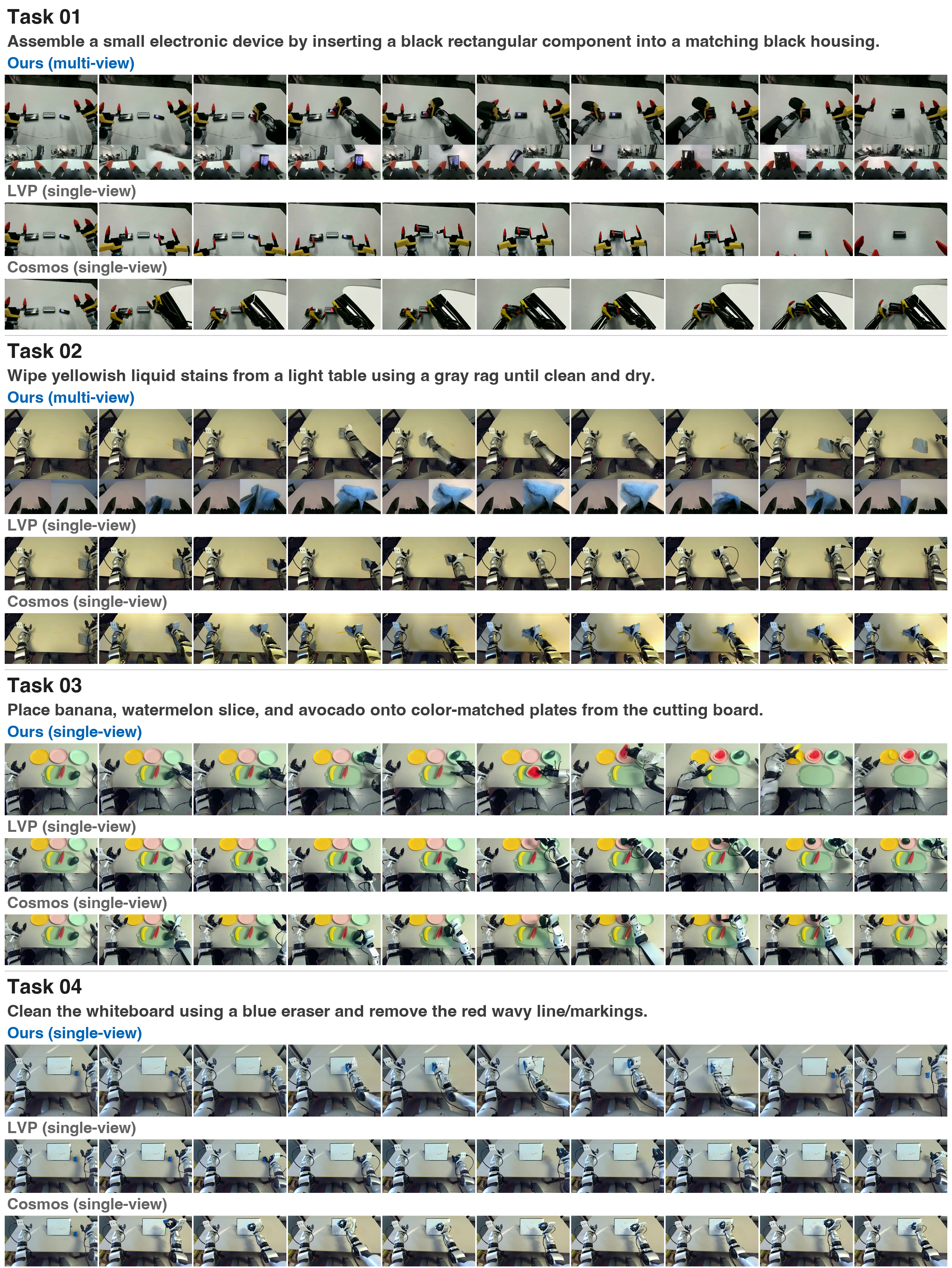}
    \vspace{-5mm}
    \caption{\textbf{Zero-shot qualitative comparison on language--action alignment and multi-view coherence.}
    Side-by-side grids under identical conditioning (same initial frame(s), prompt, and camera layout), comparing \textsc{Qwen-RobotWorld} against LVP and Cosmos2.5-14B.}
    \label{fig:lamv_zero_shot}
    \vspace{-6mm}
\end{figure}

Building on the single-model capabilities demonstrated above, we next examine whether these gains persist under controlled model-to-model comparisons.
Aggregate embodied-world-model scores can entangle three different failure sources: instruction mismatch, cross-view inconsistency, and generic visual degradation. To isolate these factors, we perform a \textbf{zero-shot} side-by-side comparison on four Unitree G1 tasks against two strong embodied baselines, LVP and \textbf{Cosmos2.5-14B}.
Figure~\ref{fig:lamv_zero_shot} shows that \textsc{Qwen-RobotWorld} more consistently preserves language-grounded execution (correct object/action correspondence and cleaner goal completion) while maintaining coherent multi-view trajectories. The two baselines show different failure patterns. \textbf{LVP} more often produces incomplete task execution, while \textbf{Cosmos2.5-14B} tends to exhibit weaker alignment between the instruction and the generated manipulation outcomes in more complex cases.

To validate this behavior under a benchmark setting, we evaluate zero-shot performance on \textbf{RoboTwin-IF} (\textit{Instruction Following}), a newly proposed benchmark built on the RoboTwin simulator with many newly constructed complex tasks. Notably, although \textsc{Qwen-RobotWorld} mixes only a small amount of open-source RoboTwin data during training, it still shows strong zero-shot performance on RoboTwin-IF together with stable multi-view consistency across synchronized camera streams. These results suggest that the model's gains are not limited to a few qualitative examples, but generalize to more challenging unseen embodied tasks. Overall, \textsc{Qwen-RobotWorld} demonstrates stronger zero-shot robustness than prior baselines by better aligning instruction following, action realism, and cross-view coherence in a unified generation framework.

\begin{figure}[h]
    \centering
    \includegraphics[width=\linewidth]{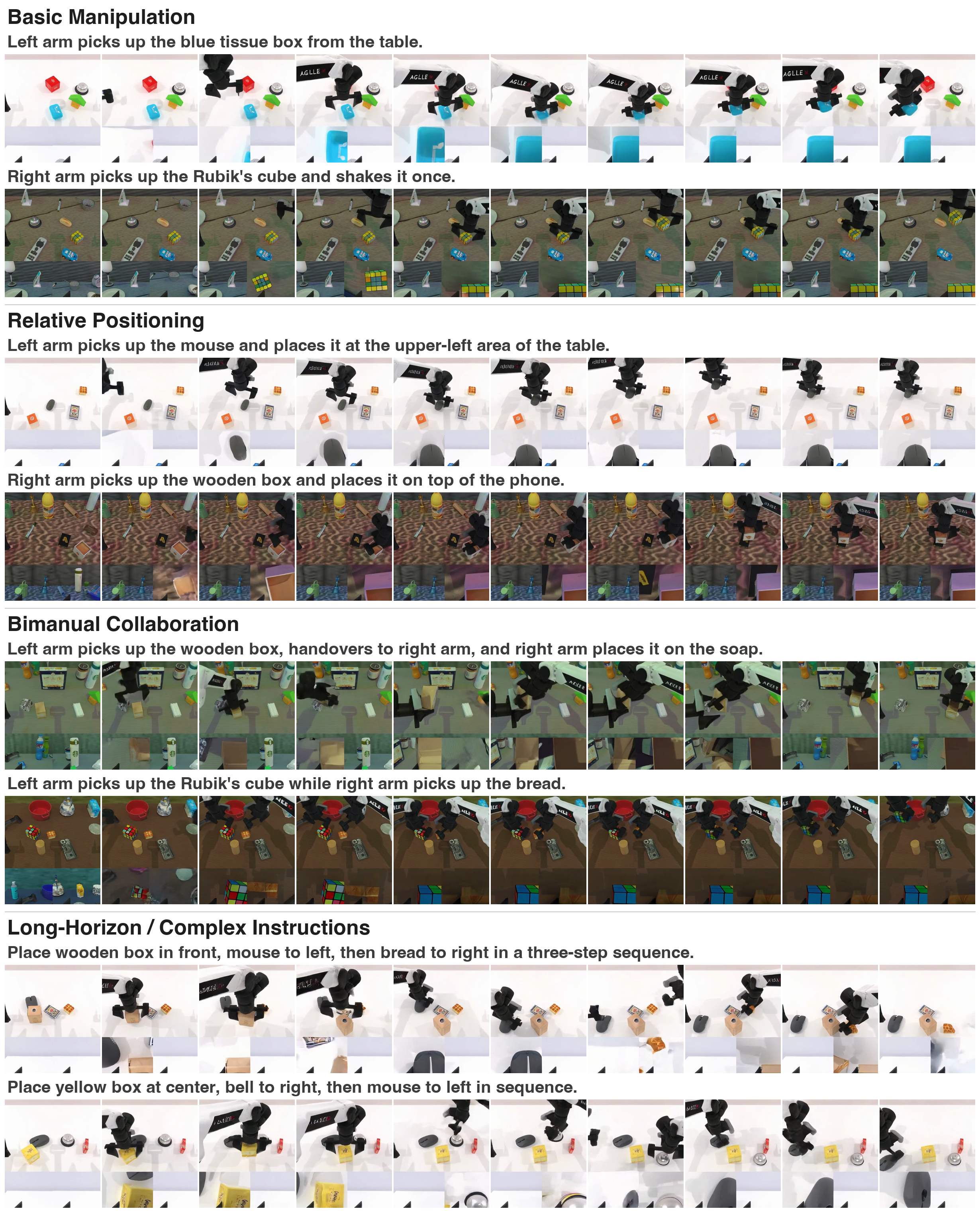}
    \caption{\textbf{RoboTwin-IF zero-shot qualitative cases.}
    The benchmark is built on the RoboTwin simulator with newly constructed complex tasks.}
    \label{fig:robotwin_zero_shot}
    \vspace{-4mm}
\end{figure}

\begin{figure}[h]
    \centering
    \includegraphics[width=\linewidth]{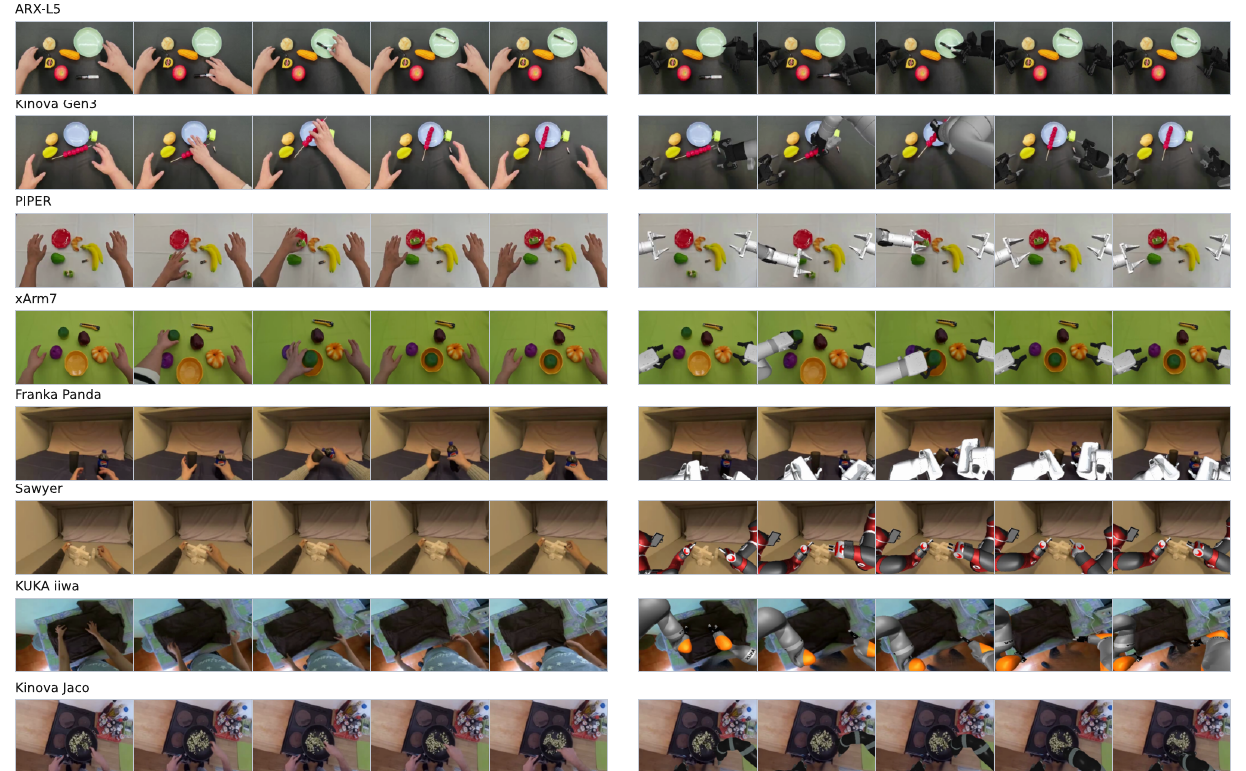}
    \caption{\textbf{Human-to-robot transfer.}
    Across eight target embodiments, each row compares a human demonstration (left) with the synthesized robot execution (right) for the same task, using five uniformly sampled frames per video.
    The generated trajectories preserve task intent while adapting motion to embodiment-specific kinematic constraints.}
    \label{fig:human2robot_gallery}
    \vspace{-4mm}
\end{figure}

\begin{figure}[h]
    \centering
    \includegraphics[width=\linewidth]{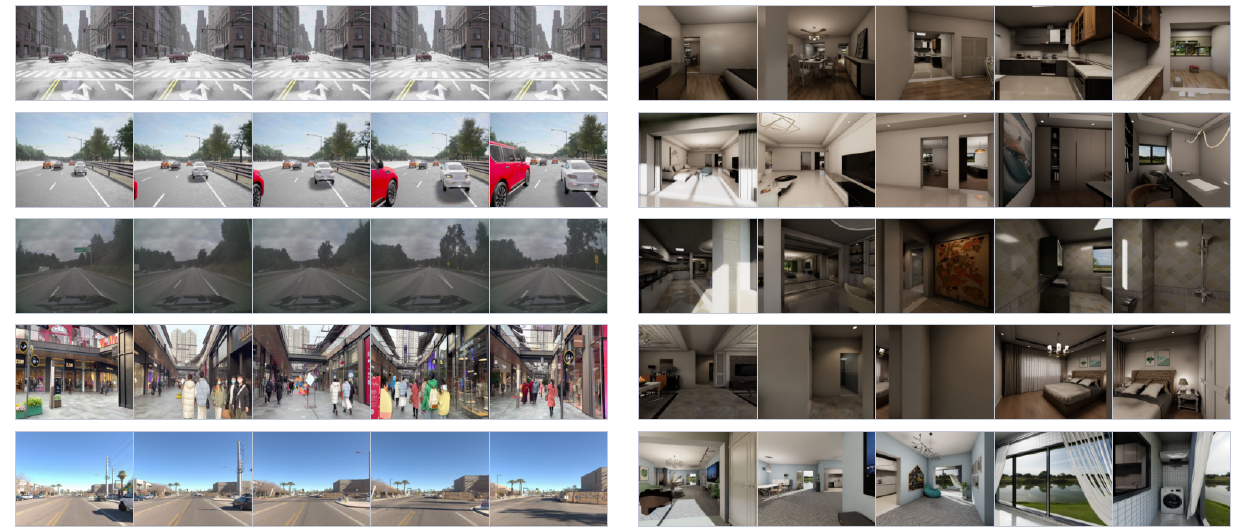}
    \caption{\textbf{Mobility generation.}
    Paired columns with five rows: \textbf{(left)~Autonomous driving} episodes from Bench2Drive, NVIDIA PhysicalAI-AD, Sekai, and Waymo; \textbf{(right)~Egocentric indoor navigation} from VLNVerse with language-guided first-person traversal.
    Each episode uses five uniformly sampled frames.}
    \label{fig:mobility_gallery}
    \vspace{-4mm}
\end{figure}

Figure~\ref{fig:robotwin_zero_shot} provides representative RoboTwin-IF zero-shot cases as qualitative evidence for this benchmark result. Each task is visualized with ten uniformly sampled frames anchored by the first and last frame, making intermediate progress and final completion directly visible. Across these newly constructed complex tasks, \textsc{Qwen-RobotWorld} preserves coherent execution and cross-view consistency, which is consistent with the quantitative RoboTwin-IF finding.

%=============================================================================
% CROSS-DOMAIN GENERALIZATION
%=============================================================================
\subsection{Cross-Domain Generalization}
\label{sec:cross_domain}

Beyond manipulation-centric evaluation, we assess the model's generalization to supplementary task families beyond the core manipulation domain.
Figure~\ref{fig:human2robot_gallery} shows human-to-robot transfer across eight target embodiments, where the model preserves task intent from human demonstrations while adapting motion to embodiment-specific kinematic constraints.
Figure~\ref{fig:mobility_gallery} covers mobility scenarios, including autonomous driving episodes from Bench2Drive, NVIDIA PhysicalAI-AD, Sekai, and Waymo, and egocentric indoor navigation episodes from VLNVerse.
Together, these results indicate that the learned language-conditioned transition model generalizes beyond a single embodiment or scenario family.

\section{Conclusion}

In this report, we present \textsc{Qwen-RobotWorld}, a language-conditioned world model framework for embodied intelligence that unifies robotic manipulation, autonomous driving, indoor navigation, and human-to-robot transfer under a shared natural language action interface. To realize this objective, we develop a three-part system: a double-stream MMDiT architecture with MLLM action encoding for semantically precise and physically grounded generation, the Embodied World Knowledge (EWK) dataset with large-scale cross-embodiment action-language alignment, and a general+expert progressive curriculum that couples broad visual priors with embodied specialization. This design enables one common backbone that can be adapted toward three representative embodied world model applications---synthetic data generation, policy evaluation, and action planning. Across both benchmark evaluations and zero-shot analyses, \textsc{Qwen-RobotWorld} demonstrates strong, consistent performance and robust multi-view instruction-following generalization. We hope this work provides a practical foundation for building embodied world models that are not only perceptually strong, but also functionally useful for downstream robotic learning and control.

% \clearpage
\appendix
% \section*{Appendix}
\setcounter{figure}{0}
\renewcommand{\thefigure}{A\arabic{figure}}

% \begin{figure}[p]
%     \centering
%     \includegraphics[width=1.5\textwidth, height=0.95\textheight, keepaspectratio]{content/showcase_gallery.pdf}
%     \caption{Showcase of Qwen-RobotWorld in embodied tasks.}
%     \label{fig:showcase_gallery}
% \end{figure}

% \begin{figure}[p]
%     \centering
%     \includegraphics[width=0.8\linewidth]{content/human2robot_gallery.pdf}
%     \caption{Human-to-robot transfer results. Each pair of rows shows 6 evenly-spaced frames sampled from the human input video (top) and the synthesized robot execution video (bottom) for the same manipulation task, across 8 target embodiments. The robot arm name is indicated at left.}
%     \label{fig:human2robot_gallery}
% \end{figure}

% \begin{figure}[p]
%     \centering
%     \includegraphics[width=0.8\linewidth]{content/mobility_gallery.pdf}
%     \caption{Autonomous driving and egocentric indoor navigation results. \textbf{Top:} 5 driving episodes sampled from Bench2Drive, NVIDIA PhysicalAI-AD, Sekai, and Waymo datasets; each row shows 7 evenly-spaced frames from a generated video. \textbf{Bottom:} 5 egocentric indoor navigation episodes from the VLNVerse evaluation set.}
%     \label{fig:mobility_gallery}
% \end{figure}

\section*{Authors}

% Contributor:
Jie Zhang\footnote{Equal contribution.}, Xiaoyue Chen\footnotemark[1], Anzhe Chen, Dayiheng Liu, Deqing Li, Gengze Zhou, Hale Yin, Haoqi Yuan, Haoyang Li, Jiahao Li, Jiazhao Zhang, Jingren Zhou, Kaiyuan Gao, Kun Yan, Lihan Jiang, Ningyuan Tang, Pei Lin, Qihang Peng, Shengming Yin, Tianhe Wu, Tianyi Yan, Xiao Xu, Yan Shu, Yanran Zhang, Ye Wang, Yi Wang, Yilei Chen, Yixian Xu, Yiyang Huang, Yuxiang Chen, Zekai Zhang, Zhendong Wang, Zixing Lei, Zhixuan Liang, Zihao Liu, Zikai Zhou, Chenxu Lv\footnotemark[2], Xiong-Hui Chen\footnote{Corresponding author.}, Chenfei Wu\footnotemark[2]

\clearpage
\bibliography{colm2024_conference}
\bibliographystyle{colm2024_conference}

\end{document}